# Four years of multi-modal odometry and mapping on the rail vehicles

Yusheng Wang[1], Weiwei Song[1, *], Yi Zhang[2], Fei Huang[2], Zhiyong Tu[1], Ruoying Li[1], Shimin Zhang[3], and Yidong Lou[1]

[1] GNSS Research Center, Wuhan University, 129 Luoyu Road, Wuhan 430079, China
[2] School of Geodesy and Geomatics, Wuhan University, 129 Luoyu Road, Wuhan 430079, China
[3] Hefei power supply section, China Railway Shanghai bureau CO., LTD, 2 Yashan Road, Hefei 230012, China

[*] Correspondence: sww@whu.edu.cn

*Abstract*—Precise, seamless, and efficient train localization as well as long-term railway environment monitoring is the essential property towards reliability, availability, maintainability, and safety (RAMS) engineering for railroad systems. Simultaneous localization and mapping (SLAM) is right at the core of solving the two problems concurrently. In this end, we propose a high-performance and versatile multi-modal framework in this paper, targeted for the odometry and mapping task for various rail vehicles. Our system is built atop an inertial-centric state estimator that tightly couples light detection and ranging (LiDAR), visual, optionally satellite navigation and map-based localization information with the convenience and extendibility of loosely coupled methods. The inertial sensors IMU and wheel encoder are treated as the primary sensor, which achieves the observations from subsystems to constrain the accelerometer and gyroscope biases. Compared to point-only LiDAR-inertial methods, our approach leverages more geometry information by introducing both track plane and electric power pillars into state estimation. The Visual-inertial subsystem also utilizes the environmental structure information by employing both lines and points. Besides, the method is capable of handling sensor failures by automatic reconfiguration bypassing failure modules. Our proposed method has been extensively tested in the long-during railway environments over four years, including general-speed, high-speed and metro, both passenger and freight traffic are investigated. Further, we aim to share, in an open way, the experience, problems, and successes of our group with the robotics community so that those that work in such environments can avoid these errors. In this view, we open source some of the datasets to benefit the research community.

## 1. Introduction

### A. Motivation

Robust multi-modal sensor integration is ideal for a wide range of robotic applications and has elicited increasing attention in the past decades. Aiming for ego-motion estimation and map-building, the simultaneous localization and mapping (SLAM) constitutes one of the most fundamental problems in robotics. Recent studies have reported successful deployment of multi-modal SLAM on handled device, unmanned ground or aerial vehicles (UGV or UAV), automated cars, even for autonomous planet exploration robots. However, the multi-modal SLAM has not been well demonstrated for railroad applications.

The existing train positioning strategy is mainly dependent on trackside infrastructures like track circuits, Balises, and axle counters. Since the accuracy of these systems is determined by the operation interval, they are neither accurate nor efficient for future intelligent rail transportation systems. Besides, these systems also require vast civil investment for construction and follow-up maintenance, which is a great burden for government.

Many railway operators have already started smart railroad practice worldwide, and the predominant solution is integration of multi-modal sensor information. For example, the project Sensors4Rail powered by Deutsche Bahn, is composed of six Light Detection and Ranging (LiDARs), four cameras and four radars. This railway project follows the idea of multi-modal SLAM, where sensors are used to locate the trains and create high-definition (HD) maps. The maps can help to analyze surroundings and monitor train journeys, where disruptions are detected and responses are triggered more quickly.

Compared with the traditional train positioning solutions, the primary improvement of multi-modal sensor integration is the significant enhancement on the localization accuracy and sensing coverage. This benefit is practically favorable for self-driving trains since we must address the precise train positions and environmental perception along the journey. Therefore, we consider multi-modal odometry and mapping for rail vehicles in this paper as visualized in Figure 1.

Our group, the Versatile and Smart Transportation group (Vast, Wuhan University), has been working on this multi-modal rail application project for more than ten years (four years for SLAM). This study has been mainly focused on precise and seamless train location, environment monitoring, communication, and autonomous driving for rail vehicles. Now all the maintenance rail vehicles in Anhui province (more than 50) are pre-installed with our GNSS/IMU/odometer/track map based localization and monitoring solution. The objective of this paper is to present the lessons learned through problems and errors encountered, decisions and successes made to the robotics research community.

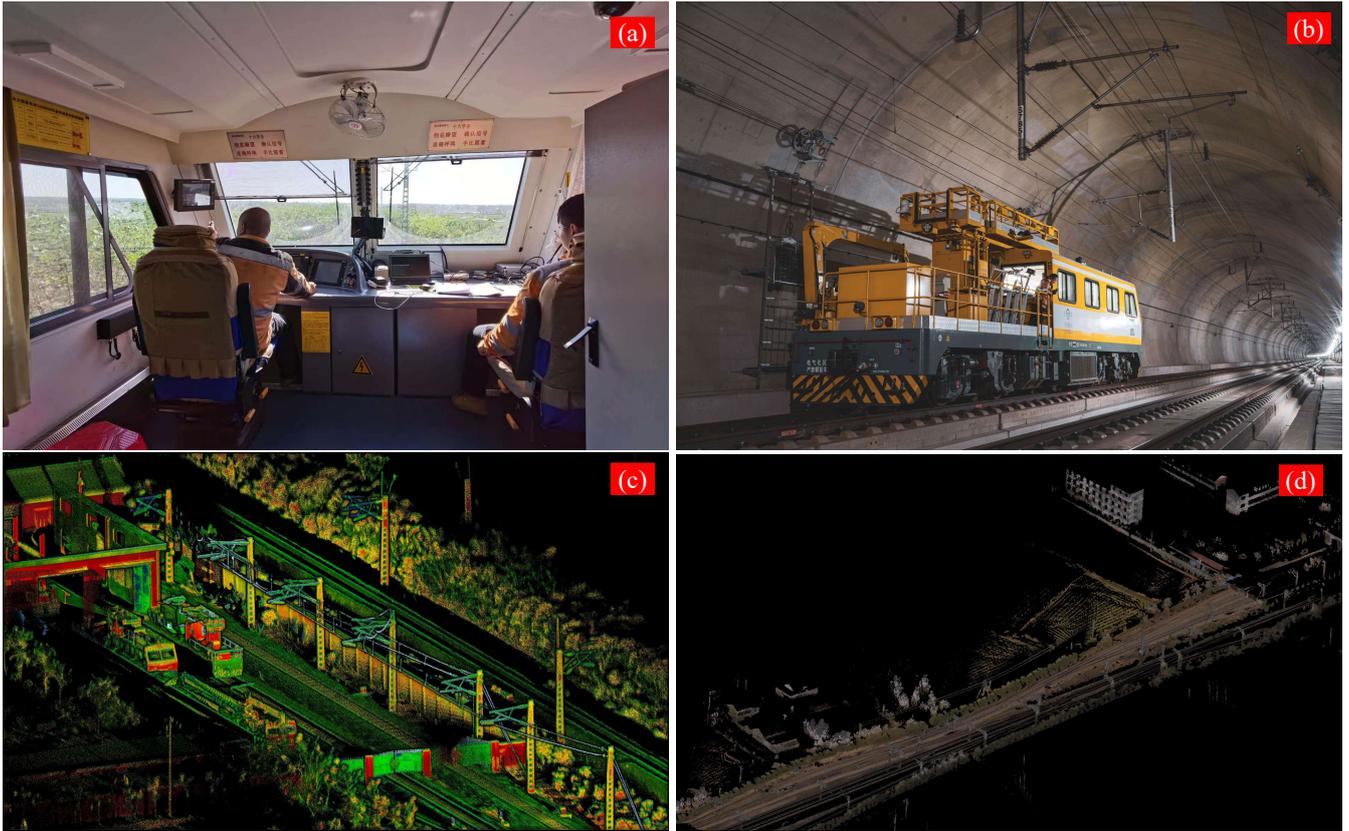

Figure 1. General view of the multi-modal SLAM on the railroad. (a) and (b) is the interior and exterior of a maintenance rail vehicle. (c) is the LiDAR mapping result of a maintenance station. (d) is the colored mapping of a railroad segment.

## *B. Challenges*

Regardless of its marvelous advantages in solving the two problems concurrently, several difficulties affect the successful application of SLAM on rail vehicles.

The predominant challenge comes from the rail vehicle long duration operation. The railway lines are usually hundreds of or thousands of kilometers long. Even the duty sector of a rail maintenance vehicle is already tens of kilometers long (Y. Wang, Song, Lou, Huang, et al., 2022). On the contrary, most of the SLAM methods are only evaluated with datasets of short distances (Xu, Cai, He, Lin, & Zhang, 2022). The operational environment of long journey is full of uncertainties, and the system robustness is in the highest priority.

For the perceptional sensors LiDARs and cameras, they can be easily affected by the environment, such as illumination variation and dust in the tunnels. Since these sensors are environment dependent, it is difficult to achieve robust performance if they are used as dominant sensors to build a SLAM system. However, most of the current SLAM system is either visual-centric (Geneva, Eckenhoff, Lee, Yang, & Huang, 2020; T. Qin, Li, & Shen, 2018) or LiDAR-centric (Shan et al., 2020; J. Zhang & Singh, 2014), which will pause or generate large temporal errors if no perceptional data is received (Tranzatto et al., 2022). In contrast, the inertial data from inertial measurement unit (IMU) or wheel encoder would not be affected by these environmental variations, but the inertial data has not been well considered. For example, IMU only constructs the preintegration factor in VINS-mono (T. Qin et al., 2018) and Lili-om (K. Li, Li, & Hanebeck, 2021), and the state estimation relies more on the visual or LiDAR information. Thus, we should fully leverage the advantage of inertial data to achieve robust positioning in large-scale environments.

The long journey also raises awareness of estimation engine design. Tightly coupled algorithms are preferred more in current multi-modal systems due to their high accuracy and robust performance. However, they may be vulnerable to cope with potential sensor failures since most of them only use a single estimation engine. Such sensor failure risks should be distributed between several estimation engines like loosely coupled ones (J. Zhang & Singh, 2018).

The railroad environment particularity also introduces many difficulties. Firstly, SLAM algorithms all have accumulated errors, where many approaches employ the revisited areas for loop closing, with drifts eliminated accordingly. However, most rail traffic has no loops due to the restriction of one-way transportation. Secondly, the rigorous safety regulations require a clearance gauge for the railway environment, where only the rail tracks, powerlines, and track side infrastructures are observable. Besides, the vegetations and buildings all should keep a certain distance away from railroads. These two regulations make most of the railroad being feature-poor districts, and may generate errors to the frame-to-frame registration process. Thirdly, the rail vehicles have a higher speed than most of the vehicles, which will cause large motion blur and less frame-to-frame correspondences. Since many

SLAM algorithms rely on the convergence of nonlinear optimization from sufficiently many iterations, insufficient feature correspondences will lead to non-convergence of optimizations. Finally, the rail vehicles are moving in a highly-constrained manner, where constant acceleration and no rotating can introduce locally unobservable IMU biases, generating ill-conditioned or even rank-deficient information matrix and result in significant scale drift for many visual SLAM approaches (Tschopp et al., 2019).

*C. Contributions*

To address the problems above, we propose an accurate and robust odometry and mapping system for railroad applications. The method treats IMU as the primary sensor and receives pose constraints from other odometry. This IMU-centric design follows a key sight: The environment free IMU state estimation becomes accurate as long as the bias can be well-constrained by other sensors. The main contributions are as follows:

1) We propose an IMU-centric SLAM framework for rail vehicles, enabling robust state estimation and large scale rail environment reconstruction. This design has the advantage to overcome potential sensor failures and a wide compatibility to different sensor setups.
2) The proposed system is self-recoverable. It fully bypasses failure modules and combines the rest odometry to handle failure cases. Unlike most of the algorithms which may fail entirely against temporary errors, our redundancy design can detect such errors and recover the individual systems.
3) We fully leverage the geometric structure of railroad environment. Track plane and pole features are extracted from raw point clouds. Besides, we extract line features and vanishing point from visual sequences.
4) The proposed system is thoroughly evaluated by real-world experiments over four years, including high-speed railways, general-speed railways, metro, covering both freight and passenger transport.
5) We open source some of the datasets to benefit the robotics research community.

*D. Organization*

The rest of the article is organized as follows. Section 2 reviews the relevant work. Section 3 gives the overview of the system. Section 4 presents some preliminary work. Section 5, 6, 7,8, and 9 presents the detailed subsystem design, followed by real-world experiments in Section 10. Finally, Section 11 concludes the article and demonstrate future research directions.

## 2. RELATED WORKS

In this section, we briefly review scholarly works on train localization and mapping as well as multi-sensory SLAM.

*A. Train Localization and Mapping*

The knowledge about the location and states of the train is of critical importance to existing automatic train control systems (ATCS), which are implemented to prevent train collisions and avoid the influence from either natural or human-made disasters.

The most usual system is based on trackside sensors, such as a Balise. This system divides the railway into separate sectors, where a Balise is placed at the beginning of each sector. When a train passes over it, the management system detects and locates that a train is within that sector. Considering its large capital investment and low localization efficiency, many researchers seek to supplement the limitations of conventional sensors with either onboard sensors or feature matching based methods. Typical onboard sensors include radio frequency identification (RFID), Doppler radar, GNSS receivers, IMU, and the tachometer or odometer. The performance of GNSS in train localization are extensively evaluated in (Acharya, Sadhu, & Ghoshal, 2011; Beugin & Marais, 2012; Q. Li & Weber, 2020; Lu & Schnieder, 2014), demonstrating that the GNSS is ideal for train localization at outdoors. Besides, the IMU and odometer can be used to compensate the long-term GNSS outages in (Q. Li & Weber, 2020). Wang et al. combined the detected loops from onboard cameras and the position estimations from millimeter radars to estimate the position of the train between the key locations (Z. Wang, Yu, Zhou, Wang, & Wu, 2021). Their experiments reveal that the infrastructure-free positioning method leads to consistent and accurate results. The feature-matching methods first establish the feature database, then the real-time positions can be acquired through matching with the database. Such features include IMU and magnetometer measurements (Heirich & Siebler, 2017a; Park & Myung, 2017), tunnel structures gathered by a 2D laser scanner (Daoust, Pomerleau, & Barfoot, 2016), and track geometry. Heirich *et.al.* sampled the IMU and magnetometer measurements and constructed a track signature, showing promising long-term stability without the use of GNSS (Heirich & Siebler, 2017b).

The potential of SLAM for rail vehicles localization and mapping has not been investigated well in literature. One of the early works, RailSLAM, jointly estimated the train state and validated the correctness of initial track map based on a general Bayesian theory (Heirich, Robertson, & Strang, 2013). The performance of Visual-inertial odometry on rail vehicles have been extensively evaluated in (Etxeberria-Garcia, Labayen, Eizaguirre, Zamalloa, & Arana-Arexolaleiba, 2021; Tschopp et al., 2019), indicating that the Visual-inertial odometry is not reliable for safety critical applications, especially at challenging scenarios with high speeds, repetitive patterns, and unfavorable illumination conditions.

*B. Multi-Sensory SLAM*

The LiDAR or visual odometry is calculated from the scan-to-scan displacement, and can be generally classified into loosely-coupled or tightly-coupled approaches. There are numerous works on loosely coupled methods in the literature (Shan & Englot, 2018; J. Zhang & Singh, 2014) due to its simplicity and extendibility, where the pose estimations from individual measurements are fused separately. On the contrary, the tightly coupled methods directly fuse isolated sensor measurements through joint optimization, and has proved advantageous for their accuracy and robust performance (K. Li et al., 2021; T. Qin et al., 2018; Xu et al., 2022).

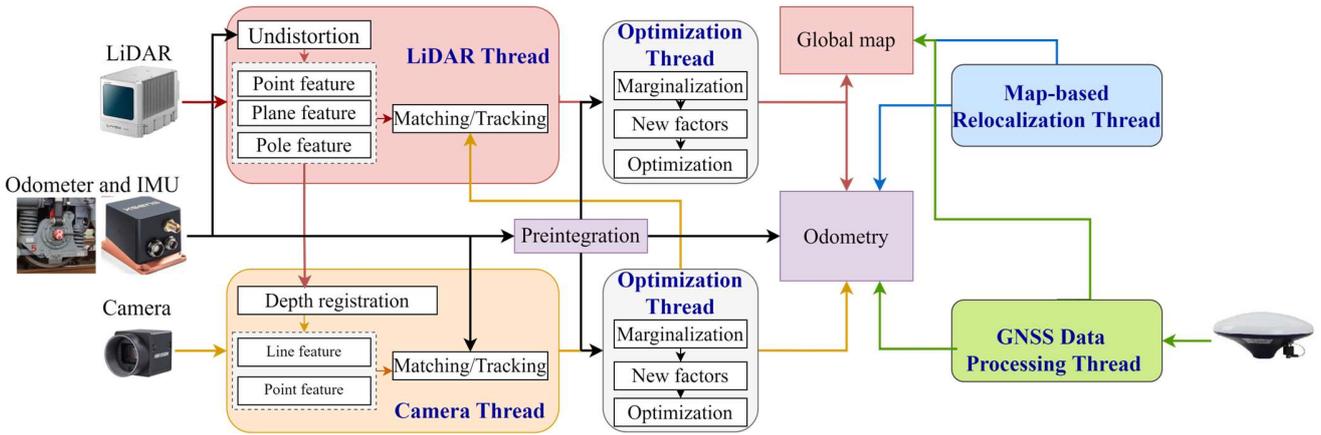

Figure 2. The overview of the proposed general framework for rail vehicle SLAM.

Many LiDAR-only SLAM are variations of point-based iterative closest point (ICP) registration (Segal, Haehnel, & Thrun, 2009). Robustness is further improved through integration with other measurements. The loosely coupled method, LOAM (J. Zhang & Singh, 2014) takes the translation and orientation calculated by the inertial measurement unit (IMU) as initial guess for registration, and discards the IMU data for further optimizations. Since the loosely-coupled systems take the odometry part as a black box and decouple the individual measurements, they are likely to suffer information loss and become less accurate. In contrast, tightly-coupled methods directly fuse isolated sensor measurements through joint optimization, and can be generally classified into optimization-based and filter-based approaches. Lio-sam (Shan et al., 2020) views IMU preintegration, LiDAR odometry and global optimization as different types of factors, and relies heavily on LiDAR odometry to further constrain pre-integrated IMU states in the factor graph. This scheme can result in loss of constraint information posed by landmarks, besides, high-frequency IMU information and precise calibration are also needed for LiDAR de-skewing. A robocentric lidar-inertial estimator LINS is proposed in (C. Qin et al., 2020), which recursively corrects the estimated state using an iterated error-state Kalman filter (ESKF). Tightly coupled methods are generally difficult to extend to other sensors and are prone to fail against sensor failures since most of them only utilize a single estimation engine. Inspired by the discussion above, we seek to propose an inertial centric SLAM system combing the accuracy and robustness advantages of tightly coupled methods with the extendibility advantage of loosely coupled methods.

To increase system robustness against sensor failures, fusion of multi-modal sensing capabilities has been explored in many scholarly works. One of the early works of LiDAR-visual-inertial SLAM, V-LOAM, is proposed in (J. Zhang & Singh, 2018), which leverages the Visual-inertial odometry as the motion model for LiDAR scan matching. Since this scheme only performs frame-to-frame motion estimation, the global consistency is not guaranteed. To cope with this problem, DV-loam (W. Wang, Liu, Wang, Luo, & Zhang, 2021) use a two-stage direct visual odometry module to estimate the coarse states, and are further refined by the LiDAR mapping module.

In LVI-SAM (Shan, Englot, Ratti, & Rus, 2021), the LiDAR-inertial and Visual-inertial subsystems can run jointly in feature-rich scenarios, or independently with detected failures in one of them. However, the "point to line" and "point-to-plane" based LiDAR odometry factor cost functions are not robust to feature-poor environments, and may generate meaningless result in metro or cave scenarios. To handle these situations, the geometry of 3D primitives are employed with the line and plane landmarks extracted in (Wisth, Camurri, Das, & Fallon, 2021).

The LiDAR or visual odometry may eventually fail due to sudden motion or observing a featureless region. The recovery mechanism has been discussed in many visual approaches (Engel, Schöps, & Cremers, 2014; Forster, Zhang, Gassner, Werlberger, & Scaramuzza, 2016; H. Huang, Lin, Liu, Zhang, & Yeung, 2020; Mur-Artal, Montiel, & Tardos, 2015). They merely depend on the visual information for temporal failure detection and use new frames for localization, which still suffers from illumination variation or degeneration. In comparison, Santamaria-Navarro, Thakker, Fan, Morrell, and Agha-mohammadi (2019) propose to use multiple parallel running odometries to detect failures. While this method is straightforward to realize, it does not utilize to its full extent the potential of all algorithms that run in parallel. Our design inherits the idea of utilizing redundant odometry to correct individual odometry failures, but we still jointly optimize the subsystems pose estimation results in the back end.

From the discussion of the literature, we can see that the multi-sensor integrated pose estimation and mapping has not been well solved and evaluated in large-scale datasets. To cope with this problem, this paper aims to achieve real-time, low-drift and robust odometry and mapping for large-scale railroad environments. The proposed method is based on our previous work where a multiple-LiDAR SLAM (Y. Wang, Song, Lou, Huang, et al., 2022) and a LiDAR-visual odometry method (Y. Wang, Song, Lou, Zhang, et al., 2022) are proposed separately. The framework is now switched to IMU-centric optimization. The LiDAR odometry now supports a great variety of LiDARs and directly extract pole features for optimization. Besides, the panoramic camera is now supported in the visual odometry. Further, the system now detects sensor failures and can recover from temporary failures.

## 3. SYSTEM OVERVIEW

We seek to estimate the trajectory and map the surrounding of a rail vehicle with multi-sensor measurements. Since the sensor setups are different for various rail vehicles, a general SLAM framework adaptive for all condition shall be developed. The overview of our system is shown in Figure 2, which is composed of five subsystems: inertial odometry, LiDAR-inertial odometry (LIO), Visual-inertial odometry (VIO), map-based localization (MO), and global navigation satellite system (GNSS) data processing (GO). Our system is formulated as a tightly coupled system, where inertial sensors IMU and odometer are viewed as the primary sensor, enabling convenient implementation and easy troubleshooting in abnormal scenarios. The constrained inertial odometry provides the prediction to the VIO and LIO. The LIO and VIO submodule extracts features from raw scans and images, which are used for state estimation. Both the two modules leverage the factor graph optimization to refine the poses. Then, the inertial odometry submodule achieves these observations to constrain the accelerometer bias and gyroscope. Besides, the accumulated inertial drifts can be corrected by MO and GO.

Before dive into details of the proposed method, we first define the notations used throughout this article in TABLE I. In addition, we define $(\cdot)_W^B$ as the transformation from world frame to the IMU frame.

TABLE I
NOTATIONS THROUGHOUT THIS PAPER

| Notations | Explanations |
|---|---|
| Coordinates | |
| $(\cdot)^W$ | The coordinates of vector $(\cdot)$ in global frame. |
| $(\cdot)^B$ | The coordinates of vector $(\cdot)$ in IMU frame. |
| $(\cdot)^L$ | The coordinates of $(\cdot)$ in LiDAR frame. |
| $(\cdot)^C$ | The coordinates of $(\cdot)$ in camera frame. |
| $(\cdot)^O$ | The coordinates of $(\cdot)$ in odometer frame. |
| Expression | |
| $(\tilde{\cdot})$ | Noisy measurement or estimation of $(\cdot)$. |
| $\otimes$ | Multiplication between two quaternions. |
| $\mathbf{p}$ | The position or translation vector. |
| $\mathbf{R}, \mathbf{q}$ | Two forms of rotation expression, $\mathbf{R} \in SO(3)$ is the rotation vector, $\mathbf{q}$ represents quaternions. |
| $x$ | The full state vector. |
| $Z$ | The full set of measurements. |
| $\eta$ | The Gaussian noise. |
| $r_{(\cdot)}$ | The calculated residual of $(\cdot)$. |
| $\delta(\cdot)$ | The estimated error of $(\cdot)$. |
| $J$ | The Jacobian matrix. |
| $e^{(\cdot)}$ | The residual of $(\cdot)$. |

## 4. METHODOLOGY

### A. Maximum-a-Posterior Problem

The train state at time $t_i$ is defined as follows:

$$x_i = [\mathbf{p}_i, \mathbf{v}_i, \mathbf{q}_i, \mathbf{b}_a, \mathbf{b}_g, \mathbf{c}_i] \quad (1)$$

where $\mathbf{p} \in \mathbb{R}^3$, $\mathbf{v} \in \mathbb{R}^3$, and $\mathbf{q} \in SO(3)$ are the position, linear velocity, and orientation vector. $\mathbf{b}_a$ and $\mathbf{b}_g$ are the usual IMU gyroscope and accelerometer biases. $\mathbf{c}$ is the scale factor of the wheel odometer.

Given the measurements $\mathcal{Z}_k$ and the history of states $\chi_k$, the maximum-a-posterior problem (MAP) problem can be formulated as:

$$\chi_k^* = \underset{\chi_k}{\mathrm{argmax}}\, p(\chi_k | \mathcal{Z}_k) \propto p(\chi_0) p((\mathcal{Z}_k | \chi_k)). \quad (2)$$

If the measurements are conditionally independent, then (2) can be solved through least squares minimization:

$$\chi^* = \underset{\chi_k}{\mathrm{argmin}} \sum_{i=1}^{k} \sum \|r_i\|^2 \quad (3)$$

where $r_i$ is the residual of the error between the predicted and measured value. We use a factor graph (Dellaert, 2012) to model this problem. A factor graph is a bipartite graph with two node types: factor nodes and variable nodes, and they are always connected by edges. As an intuitive way of formulating SLAM problem, graph-based optimization uses variable nodes to represent the poses of the vehicle at different points in time and edges correspond to the constraints between the poses. A new variable node is added to the graph when the pose displacements exceed a certain threshold, then the factor graph is optimized upon the insertion. For the sake of decreasing system memory usage and increasing computation efficiency, we employ the sliding window to keep a relative steady number of nodes in the local graph.

Given a sliding window containing $k$ keyframes, $X = [x_1^T, x_2^T, \ldots, x_k^T]^T$, we maximize the likelihood of the measurements and the optimal states can be acquired through least square minimization problem.

### B. Multi Sensor Calibration

Precise intrinsic and extrinsic calibration is crucial to every multisensory system. We first configure the extrinsic parameter of the LiDARs using EPnP algorithm (Lepetit, Moreno-Noguer, & Fua, 2009) in a calibration room with turntable and AprilTag based infrastructures. This procedure is implemented every time before onboard experiments to reduce the error caused by long-time abrasion. Besides, we leverage the continuous-time batch optimization method to perform the LiDAR-IMU calibration.

We employ the motion-based method to align the coordinates between the primary LiDAR and primary GNSS antenna. The coordinate alignment seeks to estimate the transformation by iteratively registering two sets of 3-DoF positions: $\mathbf{p}^B$ from estimator odometry and $\mathbf{p}^W$ from GNSS positioning. Since the GNSS results are in the WGS-84 geodetic coordinate system, we first transform it into local plane coordinate system $\mathbf{p}^{W_0}$ with Universal Transverse Mercator (UTM) projection. Then the corresponding LiDAR odometry and the GNSS positions can be expressed as:

$$\begin{aligned}\mathbf{P}^{W_0} &= \{\mathbf{p}_1^{W_0}, \mathbf{p}_2^{W_0}, \cdots, \mathbf{p}_n^{W_0}\}, \\ \mathbf{P}^B &= \{\mathbf{p}_1^B, \mathbf{p}_2^B, \cdots, \mathbf{p}_n^B\}\end{aligned} \quad (4)$$

where $\mathbf{P}^{W_0}$ and $\mathbf{P}^L$ are the two position sets in the local frame, and $n$ denotes the number of positions. Then the relationship between $\mathbf{p}_k^{W_0}$ and $\mathbf{p}_k^B$ can be formulated by:

$$\mathbf{p}_k^{W_0} = \mathbf{R}_B^{W_0} \mathbf{p}_k^B + \mathbf{p}_B^{W_0} \tag{5}$$

Assuming the gravity vector is estimated accurate enough, then the 6-DoF transformation can be simplified as a 4-DoF problem, including 3-DoF translation and yaw rotation $yaw_B^{W_0}$. Based thereon, we can obtain the $yaw_B^{W_0}$ from corresponding position pairs $(\mathbf{p}_k^{W_0}, \mathbf{p}_k^B)$ using cosine theorem:

$$\cos(yaw_B^{W_0}) = \frac{\mathbf{p}_k^{W_0} \cdot \mathbf{p}_k^B}{\|\mathbf{p}_k^{W_0}\| \|\mathbf{p}_k^B\|} \tag{6}$$

Then the $(\mathbf{R}_B^{W_0}, \mathbf{p}_B^{W_0})$ can be expressed as:

$$\mathbf{R}_B^{W_0} = \begin{bmatrix} \cos(yaw_B^{W_0}) & -\sin(yaw_B^{W_0}) & 0 \\ \sin(yaw_B^{W_0}) & \cos(yaw_B^{W_0}) & 0 \\ 0 & 0 & 1 \end{bmatrix},$$

$$\mathbf{p}_B^{W_0} = \frac{1}{n} \sum_{k=1}^{n} (\mathbf{p}_k^{W_0} - \mathbf{R}_B^{W_0} \mathbf{p}_k^B). \tag{7}$$

In practice, these parameters are solved automatically when the train drives out of the maintenance station with low velocity.

### C. Health Monitoring

Since sensor failures and temporal estimation failures are non-avoidable along the path, we introduce both hardware and software level verification to monitor system health.

The hardware-level verification is conducted at the data preprocessing stage, including data stream existence, frequency, and individual verification. The data stream existence test aims to find out whether the required data input exist or not. Since both our LIO and VIO has an inertial-centric design, they will not fail immediately when the LiDAR or visual sequence is not available. We degrade the respective state estimation weight to one third of original weight if the data stream is lost for more than ten seconds. Further, the respective sub-odometry will be closed when the stream is lost for more than thirty seconds. In this case, the system will send a warning in the terminal and keep on detecting that data stream. The subsystem will be recovered and reinitialized automatically if that stream is steady again. The data frequency test also follows this idea. The system set the stream with the lowest frequency as the primary input, and monitor the counts of other data within two consecutive frames continuously, e.g., the LiDAR is set as the primary input (10 Hz), and approximately ten frames of IMU input (100 Hz) should be found within two successive LiDAR scans. Once this criterion is not hold for five minutes, the system will send a warning to the user interface for a manual check, e.g., a yellow warning sign in the terminal. The individual verification is mainly for the perception sensors. We may encounter some perception degraded scenes for LiDARs, such as rains, fog in the winter morning, and dust caused by tunnel cleaning tasks. The laser beams reflected by such aerosol may generate many noise points around the LiDAR and decrease the LiDAR odometry accuracy. Therefore, we monitor the Euclidian distance of the point clouds within each scan, if 50 % of the points are below ten meters to the LiDAR, the current frame is discarded for pose estimation. Since the Visual-inertial odometry may fail against the sudden illuminance variations, we first transform the gamma-compressed RGB values to linear RGB, and compute the relative luminance of each image. The image with either too high or too low value is discarded for pose estimation due to insufficient contrast.

The software-level test is performed for parallel pose estimation modules to remove clearly wrong results. We set the maximum speed of the rail vehicle as 350 km/h, and verify whether the horizontal displacement within short time interval of each odometry is beyond this limit or not. Similarly, we leverage the maximum allowable gradient information to monitor the height estimation results (e.g., The maximum allowable gradient of high-speed railway is 20‰).

### 5. INERTIAL ODOMETRY SUBSYSTEM

For the IMU sensor, we can derive the raw accelerometer and gyroscope measurements, $\hat{\mathbf{a}}$ and $\hat{\boldsymbol{\omega}}$, through:

$$\hat{\mathbf{a}}_k = \mathbf{a}_k + \mathbf{R}_W^{B_k} \boldsymbol{g}^W + \mathbf{b}_{a_k} + \boldsymbol{\eta}_a,$$
$$\hat{\boldsymbol{\omega}}_k = \boldsymbol{\omega}_k + \mathbf{b}_{\omega_k} + \boldsymbol{\eta}_\omega, \tag{8}$$

where $\boldsymbol{\eta}_a$ and $\boldsymbol{\eta}_\omega$ are the zero-mean white Gaussian noise, with $\boldsymbol{\eta}_a \sim \mathcal{N}(\mathbf{0}, \boldsymbol{\sigma}_a^2)$, $\boldsymbol{\eta}_\omega \sim \mathcal{N}(\mathbf{0}, \boldsymbol{\sigma}_\omega^2)$. $\boldsymbol{g}^W = [0, 0, g]^T$ presents the gravity vector expressed in the world frame.

The odometer mounted on the wheel is utilized to measure the longitudinal velocity of the rail vehicle along the rails, and the model of odometer sensor is given by:

$$c_k \hat{\mathbf{v}}^O = \mathbf{v}^O + \boldsymbol{\eta}_{s^O}, \tag{9}$$

where $c_k$ denotes the scale factor of the odometer modeled as random walk, with $\boldsymbol{\eta}_{c^O} \sim \mathcal{N}(\mathbf{0}, \boldsymbol{\sigma}_{c^O}^2)$. $\mathbf{v}^O$ is the measured absolute velocity expressed in the odometer frame:

$$\mathbf{v}^O = \frac{n_{odo}}{N_{odo}} \cdot \pi \cdot d_{wheel}. \tag{10}$$

We use $n_{odo}$ to represent the number of pulses per second received and $N_{odo}$ to denote the number of pulses of a full wheel turn. Besides, $d_{wheel}$ is the wheel diameter. Then the pose estimation can be achieved through synchronously collected gyroscope and odometer output, and the displacement within two consecutive frames $k$ and $k+1$ can be given as:

$$\hat{\mathbf{p}}_{O_k}^{O_{k+1}} = \mathbf{p}_{O_k}^{O_{k+1}} + \boldsymbol{\eta}_{\mathbf{p}^O} \tag{11}$$

where $\boldsymbol{\eta}_{\mathbf{p}^O}$ is also the zero-mean white Gaussian noise. Based thereupon and the preintegration form in (T. Qin et al., 2018), we can formulate the IMU and odometer increment between $k$ and $k+1$ and integrate them in local $B_k$ as:

$$\boldsymbol{\alpha}_{B_{k+1}}^{B_k} = \iint_{t=k}^{k+1} \mathbf{R}_{B_t}^{B_k} (\hat{\mathbf{a}}_t - \mathbf{b}_{a_t} - \boldsymbol{\eta}_a) dt^2,$$

$$\beta_{B_{k+1}}^{B_k} = \int_{t=k}^{k+1} \mathbf{R}_{B_t}^{B_k}(\hat{\mathbf{a}}_t - \mathbf{b}_{a_t} - \boldsymbol{\eta}_a)dt,$$

$$\gamma_{B_{k+1}}^{B_k} = \int_{t=k}^{k+1} \frac{1}{2}\Omega(\hat{\boldsymbol{\omega}}_t - \mathbf{b}_{\omega_t} - \boldsymbol{\eta}_\omega)\gamma_{B_t}^{B_k}dt,$$

$$\alpha_{O_{k+1}}^{O_k} = \int_{t=k}^{k+1} \mathbf{R}_{O_t}^{O_k}(c_k\hat{\mathbf{v}}^O - \boldsymbol{\eta}_s o)dt, \quad (12)$$

where

$$\Omega(\boldsymbol{w}) = \begin{bmatrix} -[\boldsymbol{w}]_\times & \boldsymbol{w} \\ -\boldsymbol{w}^T & 0 \end{bmatrix},$$

$$[\boldsymbol{w}]_\times = \begin{bmatrix} 0 & -w_z & w_y \\ w_z & 0 & -w_x \\ -w_y & w_x & 0 \end{bmatrix}. \quad (13)$$

We can also transform $\boldsymbol{\alpha}_{O_{k+1}}^{O_k}$ into IMU frame $\boldsymbol{\phi}_{B_{k+1}}^{B_k}$ with the IMU-odometer extrinsic $\mathbf{R}_O^B$ through:

$$\boldsymbol{\phi}_{B_{k+1}}^{B_k} = \int_{t=k}^{k+1} \mathbf{R}_{B_t}^{B_k}\mathbf{R}_{O_t}^{B_t}(c_k\hat{\mathbf{v}}^O - \boldsymbol{\eta}_c o)dt, \quad (14)$$

Thus, the discrete form of preintegrated IMU/odometer measurements $[\hat{\boldsymbol{\alpha}}_{B_{i+1}}^{B_k}, \hat{\boldsymbol{\beta}}_{B_{i+1}}^{B_k}, \hat{\boldsymbol{\gamma}}_{B_{i+1}}^{B_k}, \hat{\boldsymbol{\phi}}_{B_{i+1}}^{B_k}]$ can be formulated by:

$$\hat{\boldsymbol{\alpha}}_{B_{i+1}}^{B_k} = \hat{\boldsymbol{\alpha}}_{B_i}^{B_k} + \hat{\boldsymbol{\beta}}_{B_i}^{B_k}\delta t + \frac{1}{2}R(\hat{\gamma}_{B_i}^{B_k})(\hat{\mathbf{a}}_i - \hat{\mathbf{b}}_{a_i})\delta t^2$$

$$\hat{\boldsymbol{\beta}}_{B_{i+1}}^{B_k} = \hat{\boldsymbol{\beta}}_{B_i}^{B_k} + R(\hat{\gamma}_{B_i}^{B_k})(\hat{\mathbf{a}}_i - \hat{\mathbf{b}}_{a_i})\delta t$$

$$\hat{\boldsymbol{\gamma}}_{B_{i+1}}^{B_k} = \hat{\boldsymbol{\gamma}}_{B_i}^{B_k} \otimes \begin{bmatrix} 1 \\ \frac{1}{2}(\hat{\boldsymbol{\omega}}_i - \hat{\mathbf{b}}_{\omega_i})\delta t \end{bmatrix}$$

$$\hat{\boldsymbol{\phi}}_{B_{i+1}}^{B_k} = \hat{\boldsymbol{\phi}}_{B_i}^{B_k} + R(\hat{\gamma}_{B_i}^{B_k})\hat{R}_{O_i}^{B_i}\hat{c}_i\hat{v}^{O_i}\delta t \quad (15)$$

We can derive the continuous-time model of error term state transition following (T. Qin et al., 2018):

$$\delta\dot{\mathbf{Z}}_{B_t}^{B_k} = \mathbf{F}_{B_t}\delta\mathbf{Z}_{B_t}^{B_k} + \mathbf{G}_{B_t}\boldsymbol{\eta}_{B_t}. \quad (16)$$

The Jacobian matrix and covariance are given by:

$$\mathbf{J}_{B_k} = \mathbf{I},$$
$$\boldsymbol{p}_{B_k} = \mathbf{0}. \quad (17)$$

Then the Jacobian matrix and covariance at $k+1$ can be recursively calculated through:

$$\mathbf{J}_{i+1} = \mathbf{F}_i\mathbf{J}_i,$$
$$\boldsymbol{p}_{i+1} = \mathbf{F}_i\boldsymbol{p}_i\mathbf{F}_i^T + \mathbf{G}_i\mathbf{Q}\mathbf{G}_i^T. \quad (18)$$

$\mathbf{Q}$ represents the continuous-time noise covariance matrix. We can hereby use the first-order approximation to define the correction model for preintegrated IMU/odometer measurements as:

$$\boldsymbol{\alpha}_{B_{k+1}}^{B_k} = \hat{\boldsymbol{\alpha}}_{B_{i+1}}^{B_k} + \mathbf{J}_{\mathbf{b}_a}^\alpha\delta\mathbf{b}_{a_k} + \mathbf{J}_{\mathbf{b}_\omega}^\alpha\delta\mathbf{b}_{\omega_k},$$

$$\boldsymbol{\beta}_{B_{k+1}}^{B_k} = \hat{\boldsymbol{\beta}}_{B_{i+1}}^{B_k} + \mathbf{J}_{\mathbf{b}_a}^\beta\delta\mathbf{b}_{a_k} + \mathbf{J}_{\mathbf{b}_\omega}^\beta\delta\mathbf{b}_{\omega_k},$$

$$\boldsymbol{\gamma}_{B_{k+1}}^{B_k} = \hat{\boldsymbol{\gamma}}_{B_{i+1}}^{B_k} \otimes \begin{bmatrix} 1 \\ \frac{1}{2}\mathbf{J}_{\mathbf{b}_\omega}^\gamma\delta\mathbf{b}_{\omega_k} \end{bmatrix},$$

$$\boldsymbol{\phi}_{B_{k+1}}^{B_k} = \hat{\boldsymbol{\phi}}_{B_{i+1}}^{B_k} + \mathbf{J}_{\mathbf{b}_\omega}^\phi\delta\mathbf{b}_{\omega_k} + \mathbf{J}_{c_{O_k}}^\phi\delta c^{O_k}. \quad (19)$$

Finally, the residual of preintegrated IMU/odometer measurements can be expressed as:

$$\mathbf{e}_k^{IMU} = \begin{bmatrix} \delta\boldsymbol{\alpha}_{B_{k+1}}^{B_k} & \delta\boldsymbol{\beta}_{B_{k+1}}^{B_k} & \delta\boldsymbol{\theta}_{B_{k+1}}^{B_k} & \delta\mathbf{b}_a & \delta\mathbf{b}_g & \delta\boldsymbol{\phi}_{B_{k+1}}^{B_k} & \delta c \end{bmatrix}^T$$

$$= \begin{bmatrix} \mathbf{R}_W^{B_k}\left(\mathbf{p}_{B_{k+1}}^W - \mathbf{p}_{B_k}^W + \frac{1}{2}\boldsymbol{g}^W\Delta t_k^2 - \mathbf{v}_{B_k}^W\Delta t_k\right) - \hat{\boldsymbol{\alpha}}_{B_{k+1}}^{B_k} \\ \mathbf{R}_W^{B_k}\left(\mathbf{v}_{B_{k+1}}^W + \boldsymbol{g}^W\Delta t_k - \mathbf{v}_{B_k}^W\right) - \hat{\boldsymbol{\beta}}_{B_{k+1}}^{B_k} \\ 2\left[\left(\mathbf{q}_{B_k}^W\right)^{-1} \otimes \left(\mathbf{q}_{B_{k+1}}^W\right) \otimes \left(\hat{\boldsymbol{\gamma}}_{B_{k+1}}^{B_k}\right)^{-1}\right]_{2:4} \\ \mathbf{b}_{a_{k+1}} - \mathbf{b}_{a_k} \\ \mathbf{b}_{g_{k+1}} - \mathbf{b}_{g_k} \\ \mathbf{R}_W^{B_k}\left(\mathbf{p}_{B_{k+1}}^W - \mathbf{p}_{B_k}^W + \mathbf{R}_{B_{k+1}}^W\mathbf{p}_{O_{k+1}}^{B_{k+1}}\right) - \hat{\boldsymbol{\phi}}_{B_{k+1}}^{B_k} \\ c_{k+1} - c_k \end{bmatrix}. \quad (20)$$

We use $\delta\boldsymbol{\theta}_{B_{k+1}}^{B_k}$ to represent the error state of a quaternion, and $[\cdot]_{2:4}$ to take out the *xyz* elements from a quaternion.

Since the VIO and LIO are both used to constrain inertial preintegration measurements, we can obtain $\mathbf{p}_{B_k}^W$, $\mathbf{v}_{B_k}^W$ and $\mathbf{q}_k^W$ from them. Besides, the bias errors are jointly optimized in the graph. With the residuals $\mathbf{e}_k^{LIO}$ and $\mathbf{e}_k^{VIO}$ calculated by LIO and VIO, the inertial odometry optimization problem can be defined by:

$$E = \sum(\mathbf{e}_k^{LIO})^T\mathbf{W}_k^{-1}\mathbf{e}_k^{LIO} + \sum(\mathbf{e}_k^{VIO})^T\mathbf{W}_k^{-1}\mathbf{e}_k^{VIO} +$$
$$\sum(\mathbf{e}_k^{MO})^T\mathbf{W}_k^{-1}\mathbf{e}_k^{MO} + \sum(\mathbf{e}_k^{GO})^T\mathbf{W}_k^{-1}\mathbf{e}_k^{GO} + E_k \quad (21)$$

where $\mathbf{W}_k$ is the covariance matrix and $E_k$ is the marginalized prior by Schur-complement. Since both the pose estimation of LIO and VIO are not globally referenced, we only utilize the relative state estimation as local constraints to correct the bias of IMU preintegration. As the prior maps and GNSS measurements are not always achievable along the railroad, we only add the $\mathbf{e}_k^{MO}$ and $\mathbf{e}_k^{GO}$ when reliable information received.

## 6. LIO Subsystem

A great variety of LiDARs are included in our system, such as the mechanical spinning Velodyne VLP-16, Ouster OS1-64, and the hybrid solid-state LiDAR Livox Horizon, Avia, Tele-15, Mid-70, as well as Innovusion Jaguar Prime. We hereby develop a general framework that is suitable to both mechanical spinning and hybrid solid-state LiDARs. The too close points from LiDAR sensor are removed at first, followed by irregularities filtering and outlier removal. Then we apply the IMU/odometer increment model to correct LiDAR point motion distortion with linear interpolation.

To extract the edge and planar features, we follow the LOAM through calculating local smoothness:

$$c = \frac{1}{|S| \cdot \|\mathbf{p}_k^i\|} \left\| \sum_{j \in S, j \neq i} (\mathbf{p}_k^i - \mathbf{p}_k^j) \right\|, \quad (22)$$

where $S$ contains the set of continuous points in the local subset. The LiDAR odometry seeks to estimate the sensor motion between two consecutive scans, which is solved by performing point-to-line and point-to-plane scan matching. For each edge point $\mathbf{p}_i^\varepsilon$ in the current scan, we search two points from its nearest neighbor in the local map, $\mathbf{p}_1^\varepsilon$ and $\mathbf{p}_2^\varepsilon$. In order to increase the searching efficiency, we build local maps for edge and planar feature. The point-to-line residual $d_{\varepsilon 2\varepsilon}$ can then be formulated by:

$$d_{\varepsilon 2\varepsilon} = \frac{|(\mathbf{p}_i^\varepsilon - \mathbf{p}_1^\varepsilon) \times (\mathbf{p}_i^\varepsilon - \mathbf{p}_2^\varepsilon)|}{|\mathbf{p}_1^\varepsilon - \mathbf{p}_2^\varepsilon|}. \quad (23)$$

Similarly, for each planar point $\mathbf{p}_i^\rho$, we search for 3 nearest points from the local planar feature map $\mathbf{p}_1^\rho$, $\mathbf{p}_2^\rho$, and $\mathbf{p}_3^\rho$. The point-to-plane residual $d_{\rho 2\rho}$ is established by:

$$d_{\rho 2\rho} = \frac{\left|(\mathbf{p}_i^\rho - \mathbf{p}_1^\rho)^T \left((\mathbf{p}_1^\rho - \mathbf{p}_2^\rho) \times (\mathbf{p}_1^\rho - \mathbf{p}_3^\rho)\right)\right|}{|(\mathbf{p}_1^\rho - \mathbf{p}_2^\rho) \times (\mathbf{p}_1^\rho - \mathbf{p}_3^\rho)|}. \quad (24)$$

Suppose the number of edge and planar correspondences is $N_\varepsilon$ and $N_\rho$ in the current frame, the residual can be calculated using:

$$e_k^{LIO_{li \to li}} = \sum_{i=1}^{N_\varepsilon} (d_{\varepsilon 2\varepsilon}^i)^2 + \sum_{j=1}^{N_\rho} (d_{\rho 2\rho}^j)^2. \quad (25)$$

Considering the multi-LiDAR setup on some rail vehicles, we introduce three favor factors to ensure an accurate and robust pose estimation, and the overall residual is formulated by:

$$e_k^{LIO_{li \to li}} = \sum_{i=1}^{N_L} \frac{1}{\kappa_f \cdot \kappa_d \cdot \kappa_p} e_k^{LIO_{li \to li_i}}, \quad (26)$$

where $\kappa_f$, $\kappa_d$ and $\kappa_p$ denotes failure detection factor, degeneracy factor, and pose estimation factor, respectively.

Since the number of extracted features decreases greatly in feature-less areas, which may lead to optimization failure. We first obtain the mean value of edge and planar points in feature-rich areas (stations, crossings, and urban scenes), and set 10% of the average as the threshold $[e_\varepsilon, e_\rho]$. The value of $\kappa_f$ is defined following:

$$\kappa_f = \begin{cases} 1, & N_\varepsilon > e_\varepsilon \text{ and } N_\rho > e_\rho \\ 50, & \text{other conditions} \\ 100, & N_\varepsilon < e_\varepsilon \text{ and } N_\rho < e_\rho \end{cases}. \quad (27)$$

Since a poor geometric distribution of the feature points will lead to large estimation errors, we employ the degeneracy factor to determine the geometric degeneration. According to Zhang (J. Zhang, Kaess, & Singh, 2016), the degeneracy factor $\lambda$ reveals whether the optimization-based problems are well-conditioned or not, and can be derived from the smallest eigenvalue of the information matrix. The threshold $e_\lambda$ can be determined from the mid-point of the margin between the distribution of $\lambda$ in both well-conditioned scenes and degenerated scenes. We can hereby define $\kappa_d$ as:

$$\kappa_d = \begin{cases} 10, & \lambda \leq e_\lambda \\ 1, & \lambda > e_\lambda \end{cases}. \quad (28)$$

Since most of train motion are constant without aggressive rotation and acceleration, we use the accurate short-term IMU increment as the reference to calculate the pose estimation factor. For two consecutive frame k and k+1, the pose estimation from IMU increment and LiDAR odometry is defined as $\mathbf{p}_{B_k}^{B_{k+1}}$ and $\mathbf{p}_{L_k}^{L_{k+1}}$, and the $\kappa_p$ can be calculated using:

$$\kappa_p = 1 - \left( \frac{\left\|\mathbf{p}_{B_k}^{B_{k+1}}\right\| - \left\|\mathbf{p}_{L_k}^{L_{k+1}}\right\|}{\left\|\mathbf{p}_{B_k}^{B_{k+1}}\right\|} \right)^2. \quad (29)$$

We find the LiDAR-only odometry is over-sensitive to the vibrations caused by the joint of rail tracks and the rail track turnouts, where errors may appear in the pitch direction. Besides, the two rail tracks are not of the same height at turnings, and the LiDAR-only odometry will maintain this roll displacements even at following straight railways. Illustrated in (Shan & Englot, 2018), the segmented ground can constrain the roll and pitch rotation. However, the angle-based ground extraction is not robust for railways as the small height variations will be ignored by the segmentation, which will generate large vertical divergence for large-scale mapping tasks.

We hereby employ the rail track plane to provide ground constraints. We first detect the track bed area using the LiDAR sensor mounting height and angle. With the assumption of the LiDAR is centered between two rail tracks, we can set two candidate areas around the left and right rail tracks and search the points with local maximum height over the track bed. Two straight lines can then be fixed using random sample consensus (RANSAC) (Fischler & Bolles, 1981) method. Finally, we exploit the idea of region growing for further refinement. As a prevailing segmentation algorithm, region growing examines neighboring points of initial seed area and decides whether to add the point to the seed region or not. We set the initial seed area within the distance of 3 m ahead of the LiDAR, and the distance threshold of the search region to the fitted line is set to 0.07 m, which is the width of the track head. Note that we only extract the current two tracks where the train is on, and a maximum length of 20 m tracks are selected for each frame. Some examples of rail track extraction are shown in Figure 3.

We are now able to define a plane with the two sets of rail track points using RANSAC. The ground plane $\mathbf{m}$ can be parameterized by the normal direction vector $\mathbf{n}_p$ and a distance scalar $d_p$, $\mathbf{m} = [\mathbf{n}_p^T, d_p]^T$. Then the correspondence of each ground point between two consecutive scan $k$ and $k+1$ can be established by:

$$\mathbf{p}_{k+1}^L = \mathbf{T}_{L_k}^{L_{k+1}} \mathbf{p}_k^L \quad (30)$$

$$\mathbf{T}_{L_k}^{L_{k+1}} = \mathbf{T}_B^L \mathbf{T}_{B_k}^{B_{k+1}} \quad (31)$$

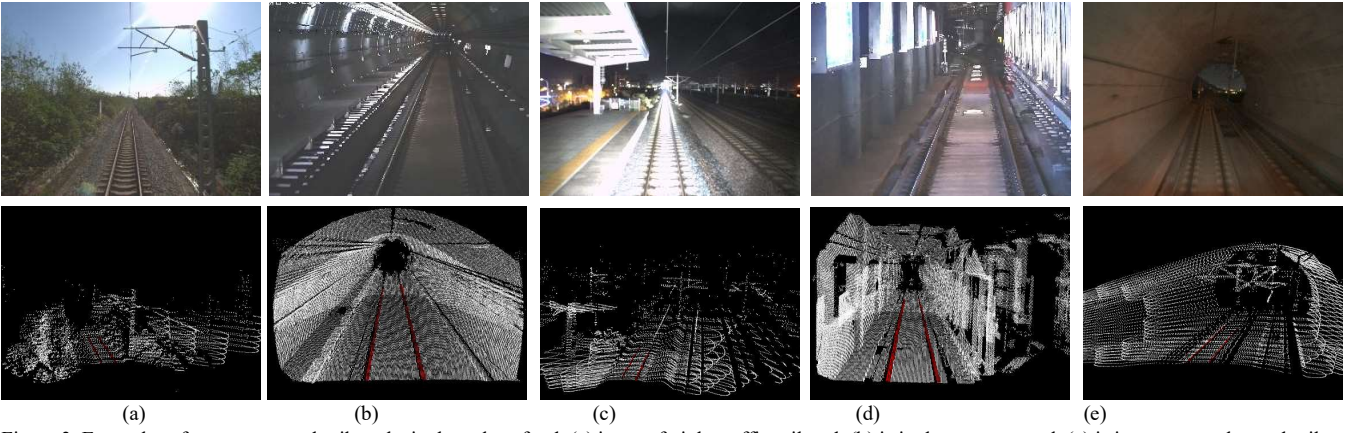

Figure 3. Examples of some extracted rail tracks in the color of red. (a) is on a freight traffic railroad. (b) is in the metro tunnel. (c) is in one general-speed railway station. (d) is in one metro station. (e) is in the high-speed railway tunnel.

where $\mathbf{p}_{k+1}^L$ and $\mathbf{p}_k^L$ is the same point expressed in frame $L_{k+1}$ and $L_k$ with the corresponding transformation defined by $\mathbf{T}_{L_k}^{L_{k+1}} = \{\mathbf{R}_{L_k}^{L_{k+1}}, \mathbf{p}_{L_k}^{L_{k+1}}\}$. Based thereon, the ground plane measurement residual can be expressed as:

$$e_k^{LIO_{pl\to pl}} = m_{k+1} - T_{L_{k+1}}^{L_k} m_k \qquad (32)$$

The longitudinal and lateral shift is hard to determine in the tunnels or in the feature-poor lanes, where mainly repetitive rail tracks and power pillars are observable. Inspired by (Kim & Kim, 2018), we aim to supplement the system with representative pole-like structures. We hereby employ the range-image based pole extractor proposed in (Dong, Chen, & Stachniss, 2021) to extract the standing or hanging pillars visualized in Figure 4. Then the mean of $x$ and $y$ marked as pole feature points can be used to optimize the translation in the lateral and longitudinal directions of the train, through minimizing the horizontal distance between two consecutive frames $k$ and $k+1$:

$$e_k^{LIO_{po\to po}} = \sum_{i=1}^{N_\varepsilon} \|\mathbf{p}_i^{L_{k+1}} - \mathbf{p}_i^{L_k}\| \qquad (33)$$

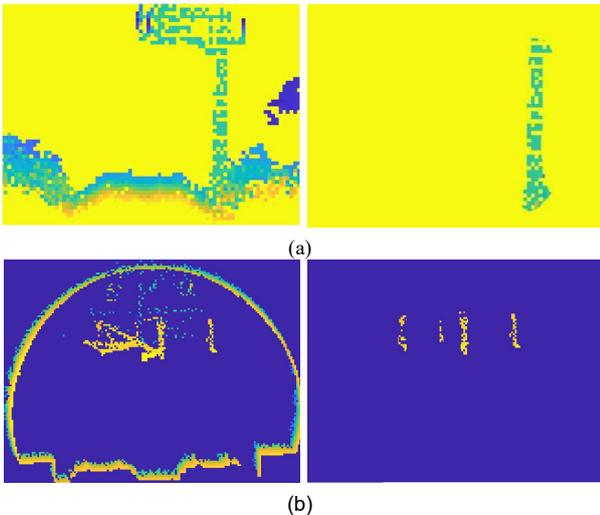

Figure 4. Illustration of the extraction of standing power pillars in (a), and the extraction of hanging pillars in (b).

To avoid the potential pillar mismatching, we first compare the $x$ coordinate of the current pole feature $x_{curr}$ with the previous $x_{pre}$. If $x_{curr}$ is much larger than $x_{pre}$, then the current translation estimation is discarded and restarted in the next frame.

Given (21), (26), (32) and (33), the minimization problem of LIO can be expressed as follows:

$$\min\left\{\sum_{i\in N_L} e_i^{LIO_{li\to li}} + \sum_{i\in N_{pl}} e_i^{LIO_{pl\to pl}} + \sum_{i\in N_{pe}} e_i^{LIO_{po\to po}} + \sum_{i\in N_I} e_i^{IMU} + e_m^{LIO} + e_{imuodom}^{prior}\right\}. \qquad (34)$$

where $N_L$, $N_{pl}$, $N_{pe}$, and $N_I$ are the number of lidar odometry residuals, plane-to-plane correspondences, pole feature correspondences and inertial preintegration factors. $e_m^{LIO}$ is the marginalization factors and $e_{imuodom}^{prior}$ is the predicted pose constraints from inertial odometry.

## 7. VIO SUBSYSTEM

For the monocular cameras, our VIO system follows the pipeline of Vins-mono (T. Qin et al., 2018). The point features are detected by (Shi, 1994), tracked by KLT sparse optical flow (Lucas & Kanade, 1981), and refined by RANSAC. For the line features extracted in monocular cameras, we employ the LSD (Von Gioi, Jakubowicz, Morel, & Randall, 2008) for line segment extraction In addition, we register LiDAR frames to the camera frames, and project the 3D distortion-free point cloud to the 2D image. Some of the LSD and depth are visualized in Figure 5. Then we can use the depth information for scale correction and joint graph optimization following (S.-S. Huang, Ma, Mu, Fu, & Hu, 2020).

$$\min\left\{\sum_{i\in N_{li}} e_i^{VIO_{li\to li}} + \sum_{i\in N_{po}} e_i^{VIO_{po\to po}} + \sum_{i\in N_I} e_i^{IMU} + e_m^{VI} + e_{imuodom}^{prior}\right\} \qquad (35)$$

where $e_i^{VIO_{li\to li}}$, $e_i^{VIO_{po\to po}}$, and $e_m^{VIO}$ are the relative line and point reprojection error, and marginalization factors. $N_{li}$ and $N_{po}$ represent the amount of line and point features tracked by other frames.

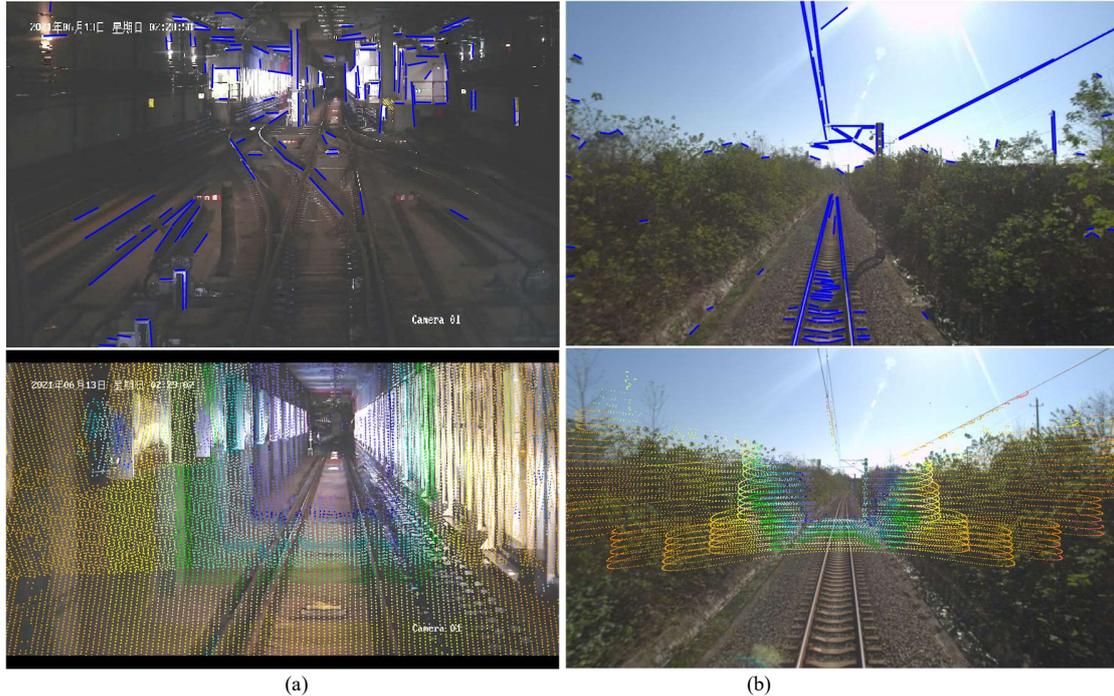

(a)                  (b)

Figure 5. Visual illustration of the extracted LSD lines and associated depth in metro environments (Innovusion Jaguar Prime LiDAR) in (a) and freight traffic railway environments (Livox Horizon LiDAR) in (b).

For the panoramic camera, our VIO system is implemented from our previous work (Y. Zhang & Huang, 2021) and performs depth enhanced direct visual SLAM similar with (W. Wang et al., 2021).

### 8. MAP MATCHING AND GNSS PROCESSING SUBSYSTEM

#### A. Map matching subsystem

We downsize the multi-modal mapping results and store them on the server. Some rail vehicles are equipped with 4G-based communication unit, which can automatically download the map segments needed for localization. When the rail vehicle enters a district with prior maps, the map-matching subsystem is awakened to match the real-time scan to the previous constructed map. To enable the real-time performance, we employ a multi-threaded normal distribution transform (NDT) method (Koide, Miura, & Menegatti, 2019) for map-based localization. NDT divides the 3D space into small cells, and calculate the local probability density function (PDF) in each cell. Then the point-to-distribution correspondences are computed within a scan pair to find the optimal transformation.

#### B. GNSS data processing subsystem

Many SLAM algorithms integrate the real-time kinematic (RTK) measurement into pose estimation to ensure high precision. However, we find that the RTK transmission channel is not always stable due to the long and complexed railroad environments, and we only use single point positioning (SPP) instead. Since the factor graph based GNSS optimization can simultaneously explore the time-correlation among historical measurements and effectively explore the time-correlation of pseudorange, carrier-phase, as well as doppler measurements. We leverage an open source GraphGNSSLib proposed in (Wen & Hsu, 2021).

### 9. MAP MANAGEMENT

The accurate scan-to-map registration of LOAM relies on the convergence of nonlinear optimization from sufficiently many iterations. However, we find the scan-to-map sometimes does not converge due to insufficient correspondences caused by large velocity. To cope with this problem, we propose a submap-based two-stage map-to-map registration, which first creates submaps based on local optimization, and utilizes the GNSS measurements (when achievable) for error correction and map registration. Once the number of iterations reaches a threshold, we introduce the GNSS positions as initial guess for ICP registration between current frame and the current accumulated submap. In addition, we leverage the GNSS information for submap-to-submap registration using the NDT. In practice, 20 keyframes are maintained in each submap, which can reduce the blurry caused by frequent correction.

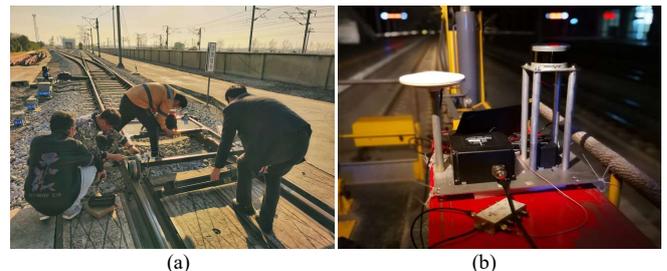

(a)                (b)

Figure 6. Visualization of the electrical railroad trolley in (a) and the detailed sensor setup in (b).

## 10. Experiment

We have conducted a series of experiments over four years with various kinds of rail vehicles mainly in four railway environments: the first are the freight traffic railways for general-speed trains, such as Fuyang-Luan line. The second are the passenger traffic railways for general-speed trains, such as Beijing-Kowloon line. The third are the passenger railways for high-speed trains, such as Hefei-Hangzhou and Hefei-Fuzhou line. The fourth are the metro railways, such as Hefei metro line 1 and metro line 2. According to the safety principle on the railroad, we can only carry out the former two experiments in the daytime, and the latter two experiments in the midnight.

Our algorithms are implemented in C++ and executed either in Ubuntu Linux using the ROS (Quigley et al., 2009) or in the Windows. Since the built-in transmission control protocol (TCP) in ROS is prone to suffer from message loss due to filled buffers. We leverage the lightweight communication and marshalling (LCM) library (A. S. Huang, Olson, & Moore, 2010) for data exchanging, which minimizes system latency and maintains high bandwidth as well.

### 10.1 Tests with Railroad Trolley

We first validate the LiDAR-inertial mapping and odometry on an electrical railroad trolley visualized in Figure 6(a). The trolley can run up to 10 km/h autonomously. One Velodyne VLP-16 and one MAPSTNAV POS1100 integrated navigation unit are attached to the trolley as shown in Figure 6(b).

The experiments were conducted in March, 2018. We use a laptop computer with a hexa-core processor i7-8750H, in a Windows operation system. The algorithm reads the point cloud information and inertial measurement from raw files. Since the LiDAR and the integrated navigation unit are hardware synchronized, we can directly find the related LiDAR and IMU frames through timestamp matching.

### 10.1.1 Accuracy tests

We manually drive the trolley on different railroads, and we employ two sequences, tro_seq1 and tro_seq2, to evaluate the localization and mapping performance of our proposed method. Both the two sequences are collected in the midnight. The first sequence tro_seq1 is a 980 m journey within 378 s and the second sequence tro_seq2 is a 1750 m journey of 680 s. The mapping results of the two sequences are visualized in Figure 7, in which the consistent and clear power lines indicating our method is of high precision even without global correction. Besides, we also provide four close views in Figure 7 for the readers to inspect the local registration accuracy.

To show the superiority of our system, we employ several state-of-the-art SLAM systems for demonstration. We select a loosely coupled LOAM (J. Zhang & Singh, 2014), a ground optimized Lego-LOAM (Shan & Englot, 2018), and a tightly coupled Lio-sam (Shan et al., 2020). Since these methods are running under ROS, we first generate the required file format using the raw measurements. Besides, we deactivate the loop detection module in Lio-sam as no revisited districts exist on the path. All the approaches merely have LiDAR and IMU inputs, and the ground truth is set as the post processed results of MAPSTNAV POS1100. The post processing software works like Inertial Explorer from NovAtel, which tightly integrates the raw GNSS and inertial measurements using forward-backward smoothing. Two criteria, root mean square error (RMSE) and maximum positioning error (MAX), are reported in TABLE II.

TABLE II
RMSE AND MAX ERRORS OF VARIOUS METHODS FOR THE TROLLEY TESTS.

| | RMSE [m] / MAX [m] | | | |
|---|---|---|---|---|
| | LOAM | LeGO-LOAM | Lio-sam | Ours |
| tro_seq1 | 15.7/179.8 | 6.9/32.7 | 1.2/3.4 | **0.8/2.1** |
| tro_seq2 | 24.2/233.5 | 14.3/167.5 | 2.1/6.6 | **1.4/3.8** |

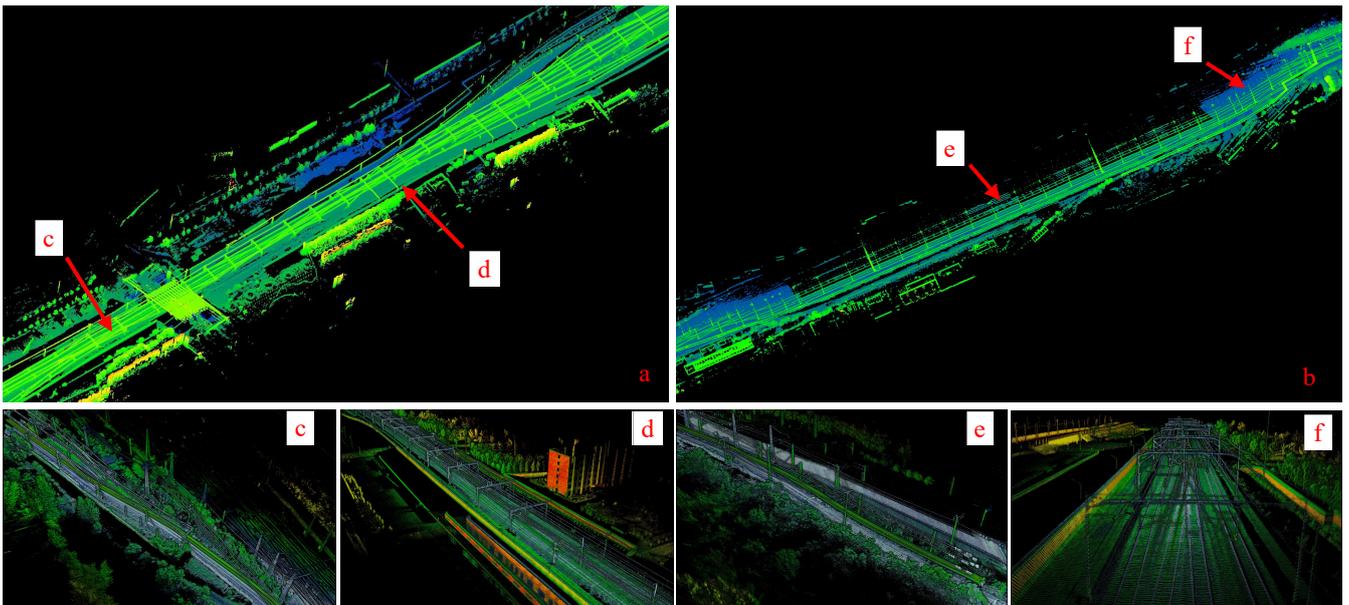

Figure 7. The mapping results of electrical trolley tests. (a) and (b) are the overall mapping of tro_seq1 and tro_seq2, the color is coded by height variations. (c) and (d) present the details of tro_seq1, the color is coded by point cloud reflectivity. Similarly, (e) and (f) are the details of tro_seq2.

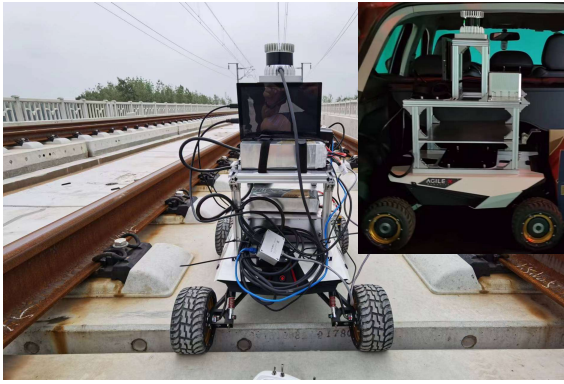

Figure 8. The work environment and sensor setup on the railroad UGV.

variations. This matches with the results where the largest error appears in the vertical direction. With gravity direction added for optimization, the tightly-coupled methods Lio-sam has a much better performance on the odometry, and it only has a 5% to 10% maximum error rate than the loosely-coupled methods.

Our method outperforms Lio-sam mainly in the vertical direction where the virtual track plane from extracted rail tracks provides strong constraint.

### 10.1.2 Lessons learned

Lesson 1: Different environments bring different challenges for SLAM algorithms. Although the well-known problems such as sharp turnings, dynamic objects or feature degeneracy does not exist for the railroad trolley test, the ground optimized SLAM should be modified considering the noisy features on the track bed. We find out that the rail tracks can provide stable ground constraints and further optimize the vertical errors.

Lesson 2: The asynchronous communication and delay between ROS nodes influence the performance and consistency of the SLAM algorithms. Unlike the selected algorithms for comparison, our approach can run under Windows operating system. The timestamp is read from observation files, ensuring an accurate time synchronization. However, the delay from Linux operating system and ROS affects the performance of the algorithm. Different computers will output non identical results, and even much worse results for computation resource limited devices. In addition, the odometry result for the same data input is not completely identical for different trials.

Since the track bed is full of crushed stones and wooden cross, LOAM extracts many unreliable edge and planar features from the point clouds. Such objects not only create many noise points in the mapping, but also influence the positioning performance. has the worst positioning and mapping performance among the selected methods. Although we can reduce the number of feature points through changing smoothness threshold, the insufficient correspondences will not converge considering the low density of VLP-16. LeGO-LOAM solves this problem using feature points clustering, the unstable clusters having less than 30 points are removed. With stones, cross, rail tracks, and track side infrastructures, the track bed is a much noisy plane compared with common roads. Therefore, the angle-based ground segmentation of LeGO-LOAM cannot distinguish such

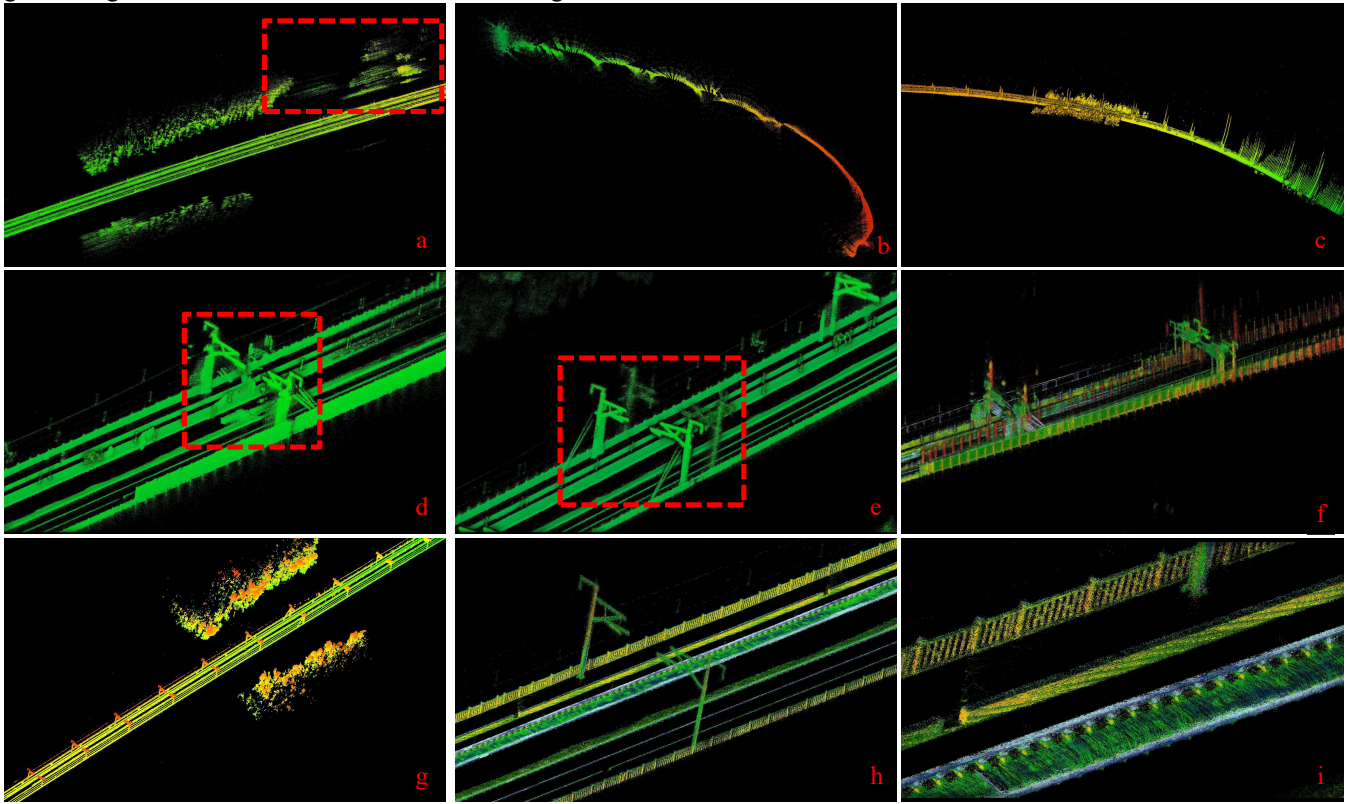

Figure 9. Visualization of some railroad UGV mapping results. (a) shows the map generated by A-LOAM, where the map is blurred in the right above corner. (b) is an example of initialization failure of Lio-sam. (c) presents the mapping of LIO-Livox. (d) and (e) show some mapping blurry or ghost points for Lili-om. (f) is the Fast-lio2 failure due to irregular vibration. (g) presents our mapping result of ugv_seq1 and the color is coded by height variations. (h) and (i) visualizes some details of our local mapping.

## 10.2 Tests with Railroad UGV

Next, we validate the LiDAR-inertial mapping and odometry on a railroad UGV as shown in Figure 8. The UGV runs on a high-speed railroad under construction to autonomously check the irregularity of rail tracks. Two LiDARs, one Ouster OS1-64 and one Livox Horizon is attached to the UGV. The inertial readings come from a Xsens MTi-680G and the built-in wheel encoders. We use a DJI Manifold 2C onboard computer to collect and process all the data. Our algorithm now runs under ROS for fast deployment and real-time processing. The localization ground truth is set as the post processing results of a MAPSTNAV M39 integrated navigation unit.

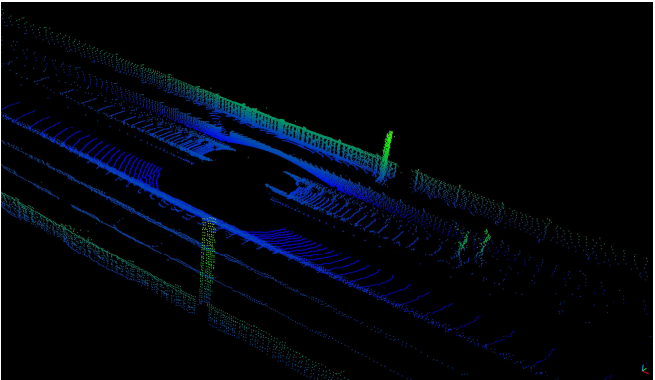

Figure 10. An example of the raw point cloud of OS1-64 LiDAR of the railroad UGV test.

### 10.2.1 Accuracy tests

Since the newly built high-speed railroad all have ballastless track as visualized in Figure 8, we can drive the railroad UGV on the track bed freely. The sequence ugv_seq is a 2.04 km journey of 699 s, many irregular motions exist along the path.

We employ several state-of-the-art SLAM systems for demonstration. We select A-LOAM, Lio-sam, two tightly-coupled system both applied for mechanical LiDARs and hybrid solid-state LiDARs, Lili-om (K. Li et al., 2021) and Fast-lio2 (Xu et al., 2022), as well as an open source tightly-coupled systems for hybrid solid-state LiDAR, LIO-Livox ("Livox-SDK/LIO-Livox: A Robust LiDAR-Inertial Odometry for Livox LiDAR," n.d.).

We first qualitatively check the system accuracy through mapping comparison. As pictured in Figure 9(a), the points in the right above corner are blurred, where the trees are not clearly visible. This is because the OS1-64 LiDAR can only see track bed and the bottom part of power pillars as shown in Figure 10, and the repetitive feature pattern will lead to partial degradation. The degraded pose estimation results further generate blurred mapping result. Since Lio-sam relies on the LiDAR odometry factor to further constrain the IMU preintegration, once the LiDAR degeneracy district is met, Lio-sam encounters large drift and generates a spinning like trajectories as visualized in Figure 9(b). Lili-om and Fast-lio2 also fails to generate clear perceptible results. In addition, Fast-lio2 is not robust to the irregular motion due to the filter-based design. When the UGV strides over the periodic barrel-drain as visualized in Figure 8, fast turning, forward and backward motion is included due to the limited wheel diameters. Fast-lio2 then generates overlapping mapping results due to inaccurate IMU predictions.

For the Livox Horizon LiDAR, the degeneration problem does not exist for all approaches. LIO-Livox relies on principal components analysis (PCA) to perform ground segmentation. However, with track fasteners and convex retaining walls, LIO-Livox fails to segment ground points properly, and generates large vertical errors. We find the patch-based feature points extraction of Lili-om has a low performance at this scene. As pictured in Figure 9(d) and Figure 9(e), many blurry or repeated features may be generated on the final map. This happens more frequently with computers of low performance. As a filter-based method, Fast-lio2 relies heavily on the data quality. Once there is a large divergence between inertial prediction and LiDAR odometry result, the final pose estimation of Fast-lio2 is incorrect and generate inconsistent mapping. As pictured in Figure 11, there exists many intervals between track slabs, and the IMU reading has a large vibration. However, the LiDAR odometry is not sensitive to these changes, and the mapping result is largely blurred as shown in Figure 9(f).

We can infer that the aforementioned methods all generate large pose estimation errors. On the other hand, our mapping is of high precision both globally and locally as pictured in Figure 9(g), Figure 9(h), and Figure 9(f).

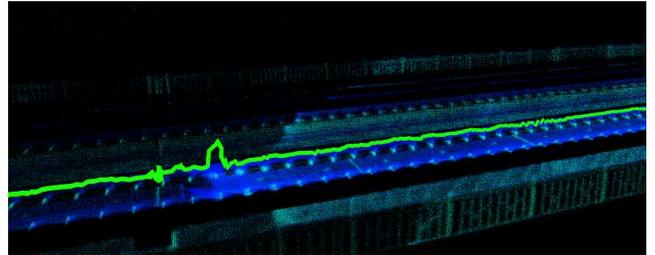

Figure 11. An example of the IMU prediction and LiDAR odometry inconsistencies. The green trajectory is the IMU prediction, where large vibration is detected within track slabs. On the other hand, the LiDAR odometry is insensitive to this vibration and retains flat.

### 10.2.2 Ablation study

We would like to further understand how the track plane and power pillars influence the system performance. To this end, we denote w/o plane and w/o pillar as our approach without track plane and power pillar constraints.

For the OS1-64 LiDAR, we plot the absolute trajectory errors (ATE) of three methods in Figure 12. We can infer that the vertical direction is the biggest error source in this scene, where our approach w/o plane has the largest errors. As for the horizontal pose errors, the curves of w/o pillar and with pillar has a similar trend. However, the maximum error is largely reduced after adding pillar constraints.

For the Livox Horizon LiDAR, we plot the ATE of three methods in Figure 13. Different from the OS1-64 LiDAR, we can directly infer that the major error source is in the horizontal direction. This is because of the limited field of view (FOV) of Livox LiDAR, where features are only observable in front of the UGV. Therefore, the pose estimation problem is only solved in the motion direction, but the errors in the heading direction cannot be estimated correctly. As pictured in Figure 14, the horizontal error is increasing continuously.

### 10.2.3 Lessons learned

Lesson 1: LiDARs of different FOV will lead to diverse pose estimation results. The spinning LiDARs of 360-degree view can generate accurate horizontal odometry, but they may have low performance at degraded districts, such as bridges and tunnels. This is because the repetitive features at the left and right side is not informative and may lead to wrong pose estimations. On the other hand, the LiDARs of limited FOV have a better performance at these scenes, since they merely focus on the right ahead direction. The distinctive objects, such as lamps, traffic signs can provide strong constraint. Since the rail application mainly focus on the distance travelled, the LiDARs with limited FOV should be preferable.

Lesson 2: Although the filter-based and optimization-based approaches are identical essentially, the filter-based approaches have a higher dependence of data quality. Adding or removing any data source implies re-designing of the filter. However, each data source is viewed as factors in optimization, we can simply add or remove any factors if needed. Besides, the filter-based approaches are more sensitive to noisy measurements, such as short period data loss, or inconsistent data sources. The noisy data sources may lead to large pose estimation errors, which is often non-reversible.

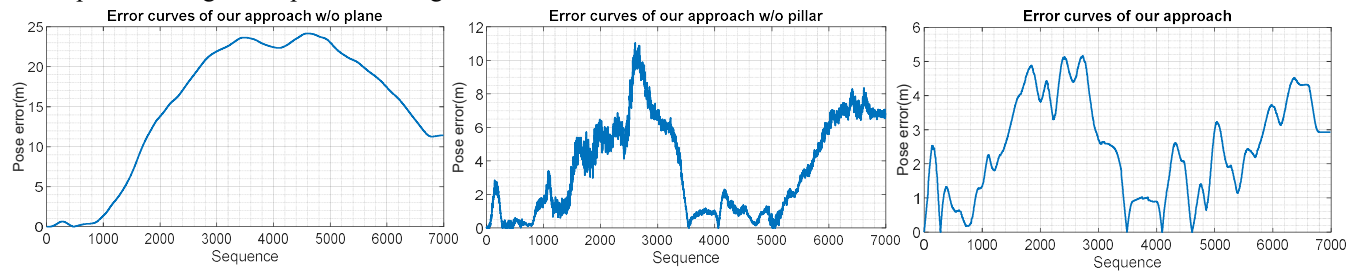

Figure 12. The three pose error curves for UGV test ablation study using OS1-64 LiDAR.

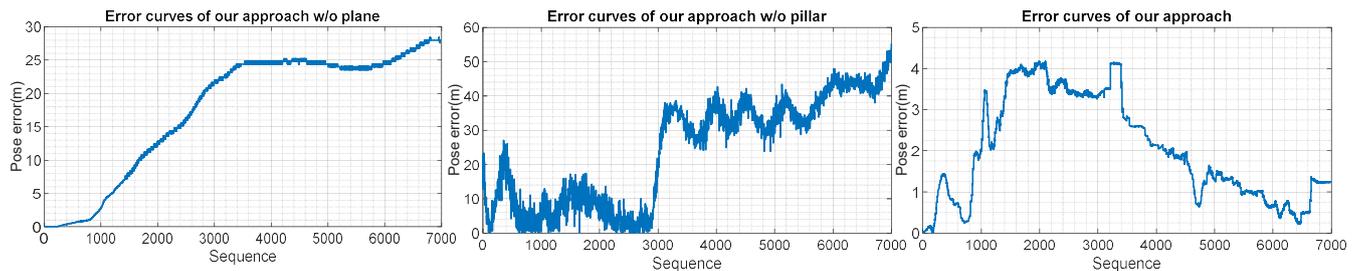

Figure 13. The three pose error curves for UGV test ablation study using Livox Horizon LiDAR.

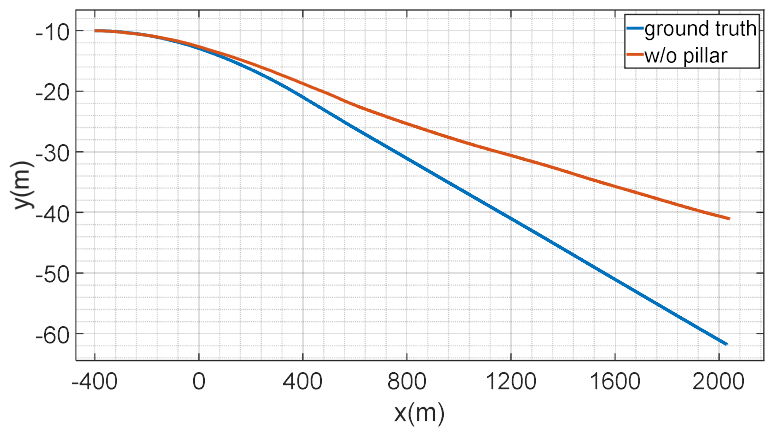

Figure 14. Visualization of the estimated trajectories of w/o pillar and the ground truth using Livox Horizon LiDAR.

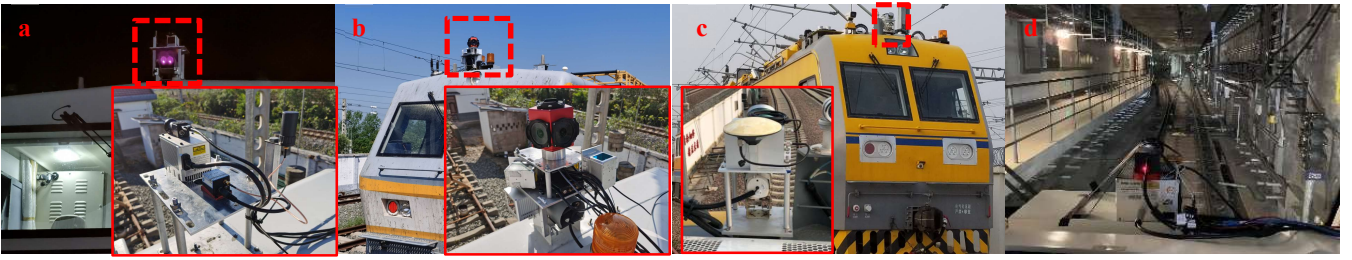

Figure 15. Sensor setups on some maintenance rail vehicles. (a) is the LiDAR-visual-inertial system setup, note that both monocular and stereo camera can be used. (b) presents an example of multi-LiDAR setup on the maintenance rail vehicles. (c) is an example of LiDAR-inertial system setup, the LiDAR can be Velodyne VLP-16 or Ouster OS1-64. (d) shows an example of LiDAR-visual-inertial system for the metro maintenance vehicles, the LiDAR used is an Innovusion Jaguar.

### 10.3 Tests with Maintenance Rail Vehicles

Further, we validate the multi-modal mapping and odometry on various railroad maintenance rail vehicles as shown in Figure 15. Following the strict safety regulations on the railroad, all the sensor platforms are welded carefully on the roof of the maintenance trains. All the sensors are hardware-synchronized with a u-blox EVK-M8T GNSS timing evaluation kit using GPS pulse per second (GPS-PPS). The data is captured and processed real-time by our customized onboard computer, with either Intel i7 7700HQ or i9 9980HK CPU, 64GB memory. The ground-truth is kept by the post-processing result of a navigation-grade IMU and a high-precision GNSS receiver, with RTK corrections sent from Qianxun SI. Note that no RTK data is included in the multi sensor system for optimization, only SPP is employed. As listed in TABLE III, eight sequences are employed for illustration.

TABLE III
DETAILS OF ALL THE SEQUENCES FOR EVALUATION

| Sequence | Line | Length (km) | Duration (s) | Sensor setup |
|---|---|---|---|---|
| General-speed passenger traffic railway | | | | |
| 201809271312 | Beijing-Kowloon | 29.4 | 1830 | VLP16 LiDAR, GNSS, IMU, odometer |
| 202106301217 | Beijing-Kowloon | 38.7 | 2752 | multiple LiDAR, panoramic camera, GNSS, IMU |
| General-speed freight traffic railway | | | | |
| 201908240653 | Fuyang-Luan | 21.9 | 1108 | VLP16 LiDAR, GNSS, IMU, odometer |
| 202006301245 | Fuyang-Luan | 14.8 | 758 | multiple LiDAR, panoramic camera, monocular camera, GNSS, IMU, |
| High-speed passenger traffic railway | | | | |
| 202109230014 | Hefei-Hangzhou | 8.4 | 872 | Horizon LiDAR, monocular camera, GNSS, IMU |
| 202110120115 | Hefei-Fuzhou | 6.3 | 684 | Avia LiDAR, monocular camera, GNSS, IMU |
| Metro railway | | | | |
| 202105240121 | Hefei Line 1 | 2.3 | 267 | Innovusion LiDAR, monocular camera, IMU |
| 202106100230 | Hefei Line 1 | 1.9 | 235 | Innovusion LiDAR, monocular camera, IMU |

TABLE IV
ACCURACY EVALUATION FOR ALL THE SEQUENCES
RMSE [m] / MAX [m], with - and bold number indicates meaningless and best result, respectively.

| | Lio-sam | Lili-om | Fast-lio2 | LIO-Livox | R2LIVE | GIO | Ours |
|---|---|---|---|---|---|---|---|
| 201809271312 | - / - | 6.81/14.88 | 3.57/12.39 | - / - | - / - | **2.67/9.48** | 3.96/14.71 |
| 202106301217 | - / - | - / - | 2.16/6.74 | 8.23/21.35 | - / - | 2.47/8.27 | **1.79/3.5** |
| 201908240653 | - / - | 3.97/11.26 | 3.57/11.55 | - / - | - / - | 2.35/7.16 | **1.59/8.62** |
| 202006301245 | - / - | - / - | 1.87/6.56 | 6.98/29.55 | - / - | 2.48/6.89 | **1.45/5.23** |
| 202109230014 | - / - | 1.76/6.8 | - / - | 7.85/32.11 | - / - | 2.29/5.35 | **1.73/4.68** |
| 202110120115 | - / - | - / - | 5.53/28.76 | 8.52/36.33 | - / - | 4.94/19.31 | **2.76/6.97** |
| 202105240121 | 11.33/69.5 | 8.7/38.76 | 7.77/21.36 | - / - | 5.17 / 25.21 | 5.34/24.88 | **4.98/23.57** |
| 202106100230 | 8.59/34.78 | 7.43/24.15 | 5.68/19.97 | - / - | 4.25/16.37 | 3.97/18.53 | **3.77/15.69** |

#### 10.3.1 Accuracy tests

We first evaluate the accuracy of our proposed multi-sensory system. Unlike the MMS-based approaches which need complicated post processing procedures, our system can generate real-time 1 Hz map while travelling. Since MMS based methods are majorly through direct-georeferencing, inaccurate position and attitude measurements will generate blurred mapping results. To show the excellence of system, we first plot some local mapping results in Figure 16. The clear enough and vivid results demonstrate that our mapping is of high precision locally. Further, we align some multiple LiDAR mapping results with satellite images in Figure 17. Since two long-range LiDARs, Livox Tele-15, are equipped on left-right side of the maintenance rail vehicle, the features even more than 300 m away from the railroads are well-matched with the satellite image. Besides, the power pillars, trees, and buildings can be easily distinguished based on the height colors.

Further, we transform some map segments into WGS-84 coordinates and directly project them onto georeferenced aerial photogrammetry images. Since each pixel on the georeferenced image is associated with global positions, we can directly measure the global mapping errors on the image as visualized in Figure 18. By comparing buildings edges on the image and point clouds, we can determine the horizontal error is less than 0.5 m. In addition, we leverage the extracted current rail tracks to find the track centers, and project rail track centers along with the extracted power pillars onto the georeferenced aerial photogrammetry image as visualized in Figure 19 to visualize the global error. It is seen the rail track centers and power pillars are well-matched with the visual map in open areas. By comparing the positions of power pillars, we can determine the error is less than 1.0 m.

We employ Lio-sam, Lili-om, Fast-lio2, LIO-Livox, and a LiDAR-visual-inertial system R2LIVE (Lin, Zheng, Xu, & Zhang, 2021) for evaluation. For the multi-LiDAR setup, the selected algorithms directly take the calibrated and merged point clouds as input. To enable a fair comparison, we manually add the same GNSS constraints to the back-end using GTSAM (Dellaert, 2012). We use GIO to represent the GNSS, IMU, and odometer tightly-coupled system developed by our group. Two criteria, RMSE and MAX, are reported in TABLE IV.

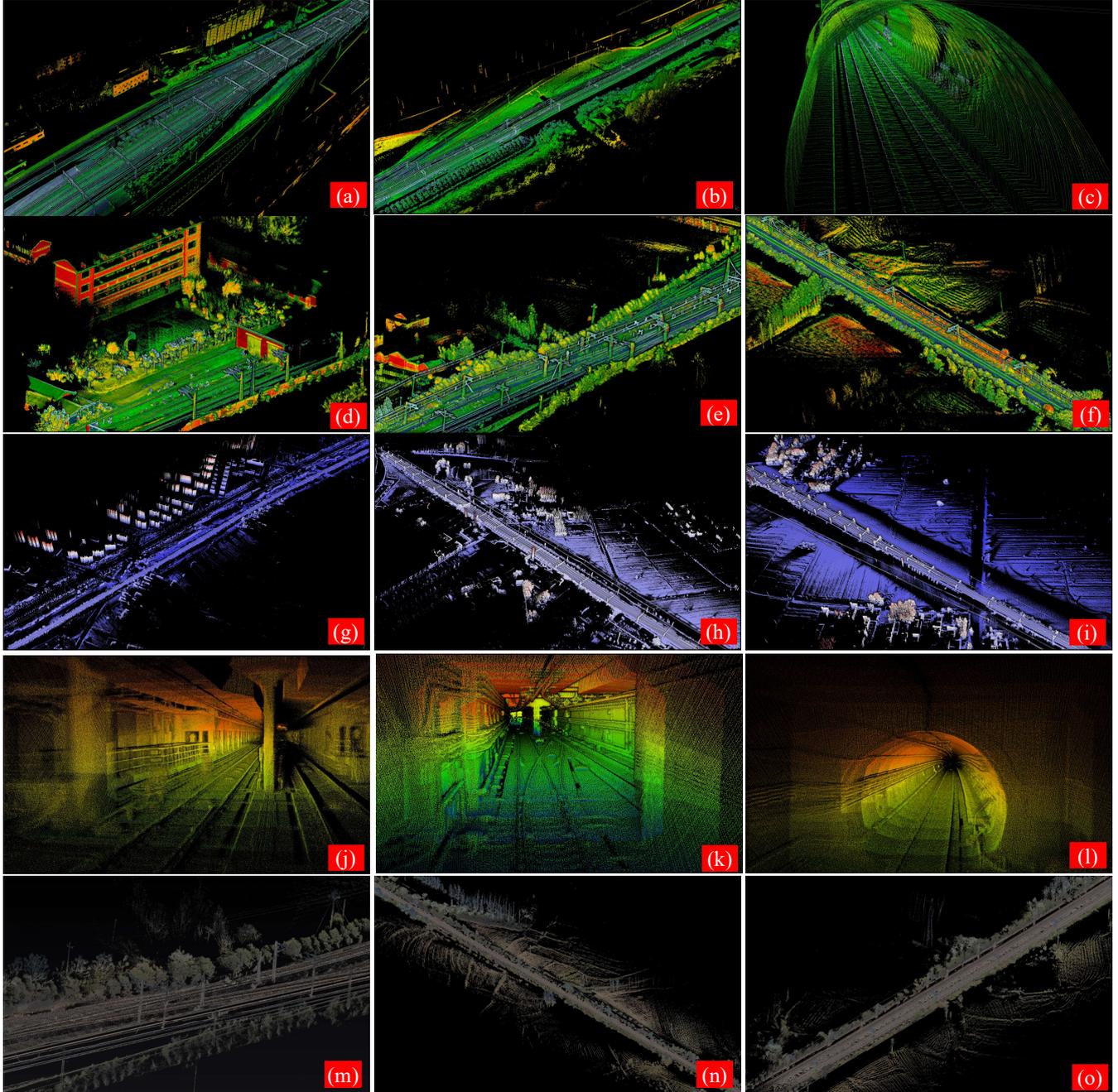

Figure 16. Mapping details of maintenance rail vehicles tests. (a), (b), and (c) are the maps generated using spinning LiDARs. (a) is near the train station, (b) shows a segment of passenger traffic railroad, and (c) is inside a tunnel. (d), (e), and (f) are the maps created using Livox series LiDARs, and the color is coded by raw reflectivity. (d) and (e) show examples of maintenance rail vehicle stations. (f) presents an example of freight traffic railroad. (g), (h), and (i) are the maps built by multiple LiDARs, and the color is represented by height variations. (g) is near a city, where the tall buildings more than 200 meters away from the railroad are clearly visible. (h) and (i) are the details on one busy passenger traffic railroad. (j), (k), and (l) are the maps generated using high-resolution LiDARs in the metro tunnels, while (j) is in a metro station, (k) and (l) are examples of metro tunnels. (m), (n), and (o) are the colored mapping results using LiDAR and cameras.

For the general-speed passenger traffic railway, there are mainly two kinds of localization and mapping solutions. The first one is visualized in Figure 15(b), composed of five mid-range Livox Horizon LiDARs and two long-range Livox Tele-15 LiDARs, and a Xsens MTI-680G outputs inertial and GNSS measurements. The panoramic camera used is a Ladybug 5+ camera. The second one is shown in Figure 15(c), including one VLP-16 LiDAR and an integrated navigation system. Since Lio-sam relies on the LiDAR odometry factor to further constrain the IMU preintegration, once the LiDAR degeneracy district is met, Lio-sam encounters large drift and generates a spinning like trajectories as visualized in Figure 20(a) for sequence *201809271312*. Considering the low point cloud density, Lili-om and our approach can only extract a small number of feature points in this scenario, and relies heavily on the GNSS positioning to correct the LiDAR odometry error caused by insufficient feature correspondences. In contrast, Fast-Lio2 employs all the point clouds for calculation, and has a higher accuracy than the above two methods. However, none of the SLAM algorithms outperform GIO due to the low point cloud density, which is harmful for the LiDAR odometry.

Lili-om divides the raw scan into patches with $6 \times 7$-point, and perform an eigendecomposition for the covariance of the 3D coordinates to find potential feature points. We find this feature extraction not appropriate for large velocity scenario. For the sequence *202106301217*, where the highest speed is 75 km/h, Lili-om fails to extract enough feature points, resulting a backward motion as shown in Figure 20(b). Our multi-sensory system outperforms others with more integrated sensors.

For the general-speed freight traffic railway, the sensor setup is similar with the two sequences of the passenger railway. It is seen that the freight sequences have a higher accuracy than that of the passenger railway. This is because the Beijing-Kowloon line is a very busy line with many trains travelling around. Once there is a train on the neighboring track, the LiDARs on that side are almost blinded as shown in Figure 21, and the system merely relies on the LiDARs on the other side for estimation.

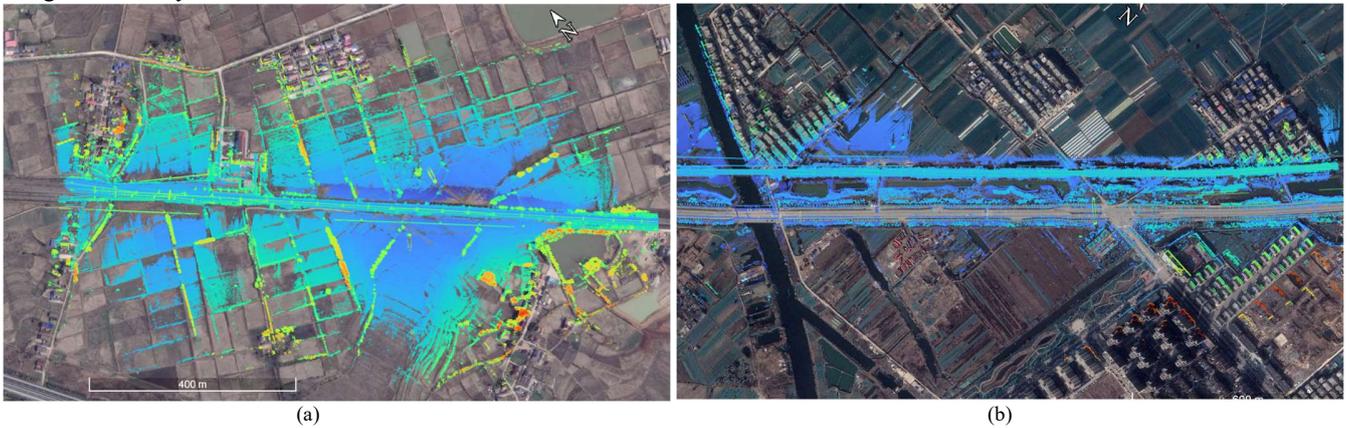

Figure 17. Two multiple LiDAR mapping results aligned with satellite image. (a) is a segment of *202006301245* and (b) is a segment of *202106301217*.

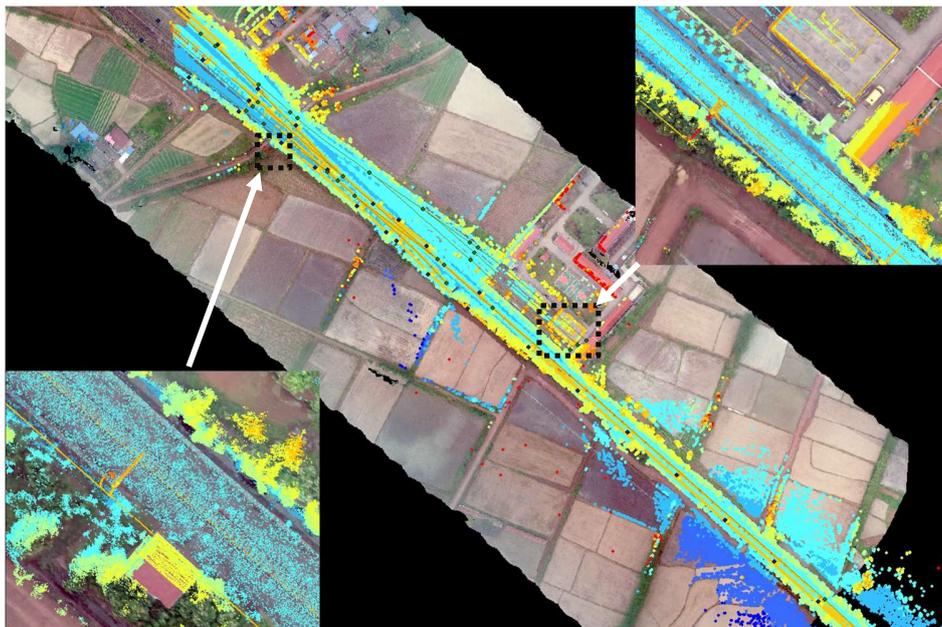

Figure 18. A segment of *202109230014* map projected onto georeferenced aerial image. The two insets are the zoom-in of detailed districts.

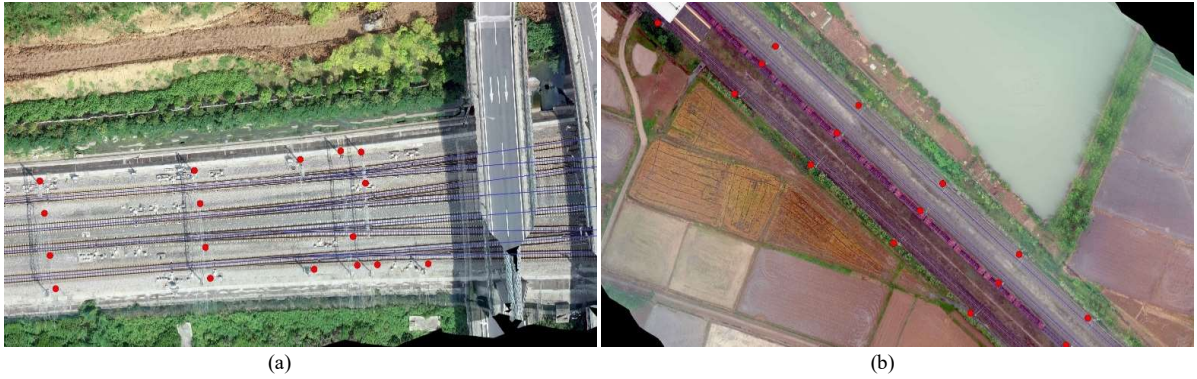

(a)                                                 (b)

Figure 19. Rail track center and power pillars extracted from multi-modal SLAM projected onto georeferenced aerial image. The red dots are the power pillars and the blue lines are the extracted track center. (a) is a segment of *201809271312* and (b) is a segment of *201908240653*.

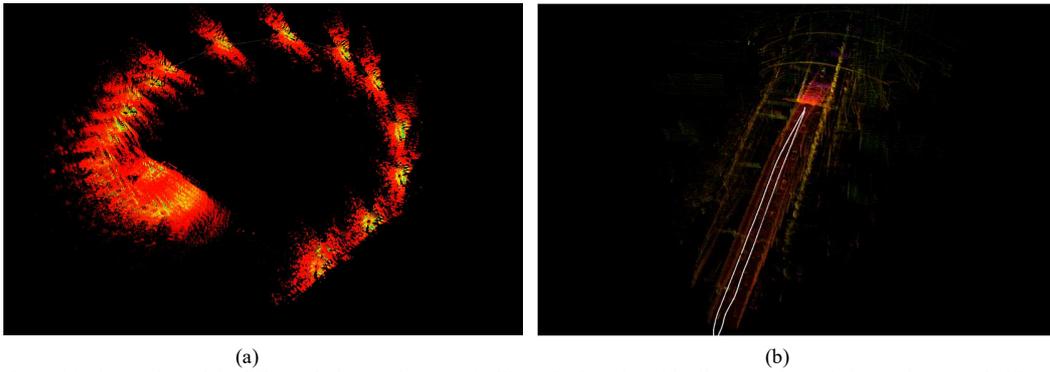

(a)                                                 (b)

Figure 20. Illustration of the failure of Lio-sam in (a) and Lili-om in (b). The white lines are the real-time trajectory of Lili-om.

Although the high-speed railway experiments are conducted in the midnight, the visual information is still applicable with the help of the train head light. The monocular camera used is a Hikvision camera illustrated in Figure 15(a). We also use a Xsens MTI-680G to provide the inertial and GNSS measurements. According to the safety regulations on the railroad, the high-speed railway maintenance rail vehicles have a slower speed than that of the general speed. It is seen that the LIO-Livox has the worst performance among all the algorithms. Firstly, LIO-Livox relies on the online LiDAR-inertial initialization process to initial the IMU noise and bias. Unlike the UGVs or unmanned aerospace vehicles (UAV) which can move freely, the dynamics of rail vehicles are highly constrained to a constant velocity. This highly constrained motion will cause insufficient axis excitement, and the initialization process is not complete. Then the LIO-Livox will turn to LiDAR-only odometry and mapping. Secondly, LIO-Livox depend on the extracted ground points to further constrain the vertical displacement. Since the Euclidean clustering is not robust to extract ground points on track bed, LIO-Livox has the largest error in the vertical direction. For the sequence *202109230014*, the maintenance vehicle is moving backwards to the station. As a filter-based tightly-coupled system, Fast-lio2 is not robust to this long-during backward motion, and generate large drift in all 6 DoF as shown in Figure 22. However, the optimization-based methods are immune to this influence. The railway scenario is not ideal for vision-based approaches, where great scale drift happens due to insufficient axis excitement. Without depth information assistance, the vision part of R2LIVE fails to provide meaningful result and ruin the whole system. Besides, when the rail vehicle starts in the turnings, the gravity vector cannot be initialized correctly as the two rail tracks are not of the same height at turning. With incorrectly initialized gravity vector, the R2LIVE will generate large errors for the sequence *202110120115*.

For the metro railway experiments, the LiDAR point cloud density has a strong influence to the positioning performance. Lio-sam fails with 16-channel Velodyne VLP-16 LiDAR even in feature-rich outdoors *201809270312* and *201908240653*. On the contrary, it survives in the feature-poor metro tunnel with only the IMU assistance, where the LiDAR is switched to a 300-channel Innovusion Jaguar Prime. As an inertial-centric solution, when all the other sensor failures are met, our system has a similar performance to the inertial-only methods. For the sequence *202105240121* and *202106100230*, when serious LiDAR and visual degeneracy problem is met, our system has an approximate accuracy to that of the GIO.

### 10.3.2 Ablation study

This experiment seeks to understand the individual accuracy of LIO and VIO module in our system, and we select sequence *202006301245* for illustration here.

For the long-during sequence *202006301245*, we plot the individual trajectory of LIO, panoramic VIO and GIO in Figure 23. It is seen that the odometry of panoramic vision SLAM is rotating around the start point. We believe this is mainly because of two reasons, the first is the bad distribution of orb features and the second is the false depth association for faraway features. Visualized in Figure 24, we set a mask on the bottom part of the panoramic image, where the static pixels on the maintenance train are discarded. The extracted orb features are mainly in the central part of the image. According to Belter et al. (Belter, Nowicki, & Skrzypczyński, 2016), the unevenly distributed features will lead to large estimation error in the unobservable directions. Although the extrinsic between the panoramic camera and LiDAR has been precisely configured, the 3D point reprojection errors will be enlarged in the distance. Therefore, the faraway orb features are associated with wrong depth. Without the assistance of GNSS positioning, our LIO also fails to provide meaningful results.

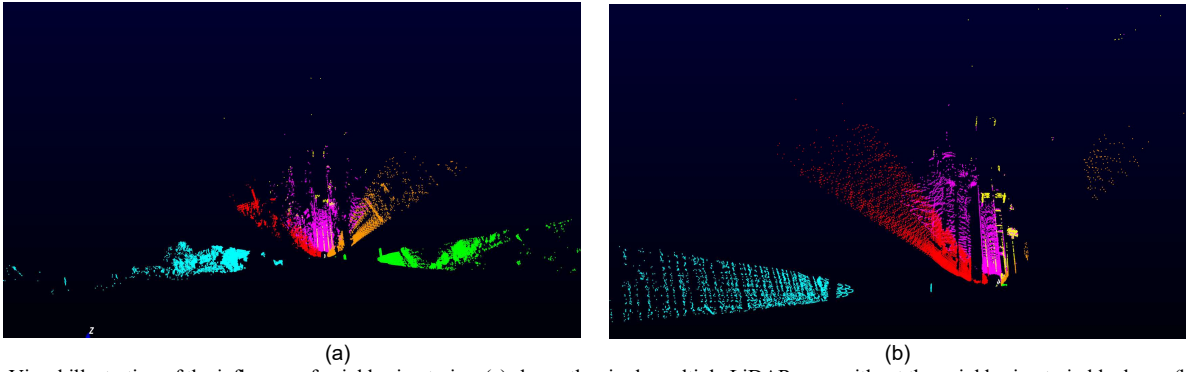

Figure 21. Visual illustration of the influence of neighboring trains. (a) shows the single multiple LiDAR scan without the neighboring train blockage. (b) presents the scan when a train is on the right side, the LiDARs on the right side are almost blinded. The color is coded by LiDAR IDs.

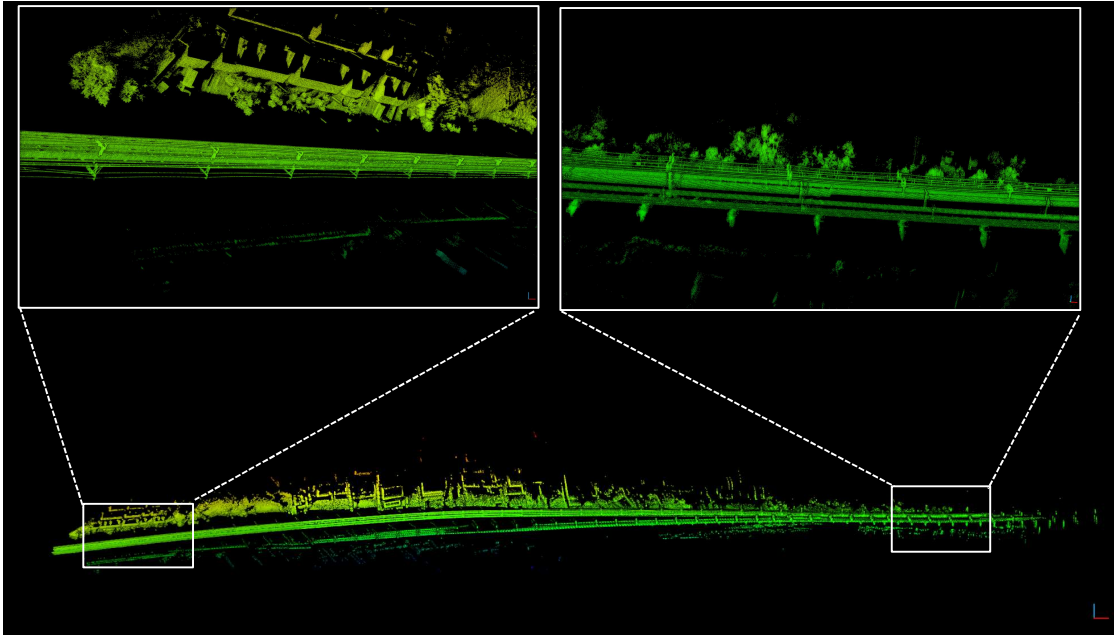

Figure 22. Visual illustration of the Fast-Lio2 failure. It is seen the tracks in the right inset is still horizontal, but the tracks in the left inset is vertical due to the long-during backward motion.

For the sequence *202110120115*, we would like to further understand the contribution of line features and vanishing points in the railroad scene. Besides, we select VINS (T. Qin et al., 2018), ROVIO (Bloesch, Omari, Hutter, & Siegwart, 2015), and MSCKF (Mourikis & Roumeliotis, 2007) for comparison. In addition, we denote VIO w/o line and VIO w/o VP as our monocular VIO without the assistance of line features and vanishing points. The respective ATE RMSE and scale errors are computed and visualized in TABLE V.

The major error source of monocular SLAM is the incorrect scale estimation. The scale information is usually obtained by initialization at the primary stage of the system, and then continuously optimized during the operation. Different from the UAVs having motion variations in all 6 DoF, the rail vehicle is moving in a very constrained motion with constant velocity. This results in insufficient axis excitement, and lead to unobservable biases. The partial unobservability of IMU biases on the limited parallel tracks results in significant scale drift. Since we have a good knowledge of the IMU biases, ROVIO can partly recover from this scene. On the other hand, VINS reinitialize multiple times along the path due to inconsistent estimator state. The scale errors at initialization stage also influence MSCKF, which diverges rapidly after initialization.

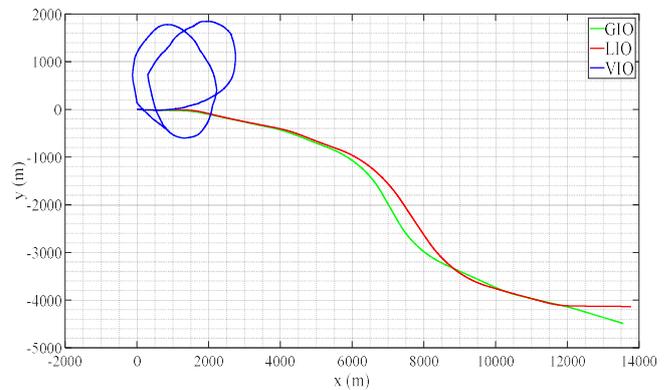

Figure 23. The trajectory comparison of sequence *202006301245*.

TABLE V
RMSE ATE(M)/SCALE ERROR EVALUATION OF MONOCULAR VISUAL ODOMETRY OF SEQUENCE *202110120115*.

|  | VINS | ROVIO | MSCKF | VIO | VIO w/o line | VIO w/o VP |
|---|---|---|---|---|---|---|
| *202006301245* | 1763.18/8.75 | 776.82/3.84 | 3927.4/19.39 | **204.8/1.49** | 698.76/6.45 | 467.13/2.37 |

We also observe that many degradation scenes exist during the trajectory. The first category is visualized in Figure 25(a) and Figure 25(b), where limited number of features points are extracted from these scenes. This inefficiency further leads to pose estimation errors. The second category is visualized in Figure 25(c) and Figure 25(d), where the repetitive and similar patterns bring difficulties to feature tracking. In these scenes, the dominant and stable line features, such as cables and tracks, can better describe the environment and has a significant pose estimation improvement as plotted in TABLE V.

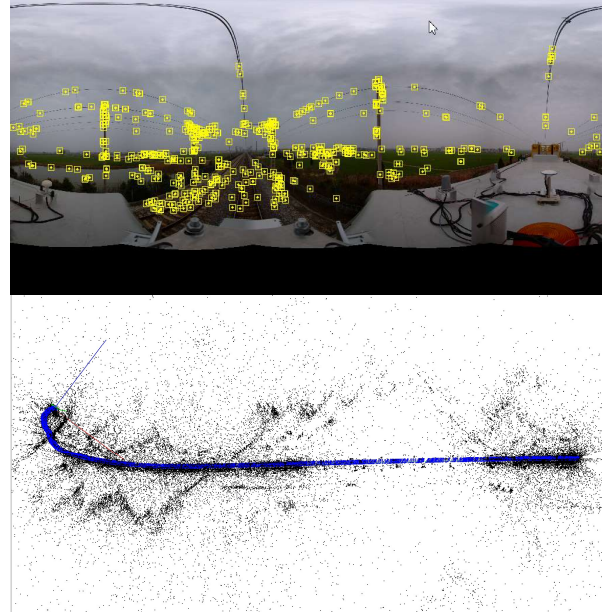

Figure 24. The feature extraction and mapping process of the panoramic SLAM in the top and bottom, respectively.

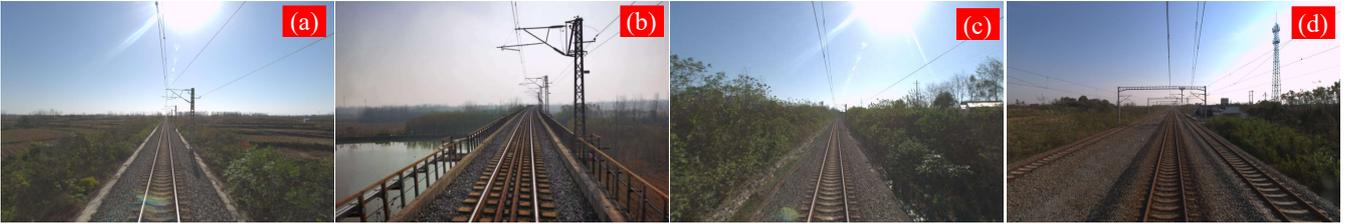

Figure 25. Examples of some degradation scenes on the railways.

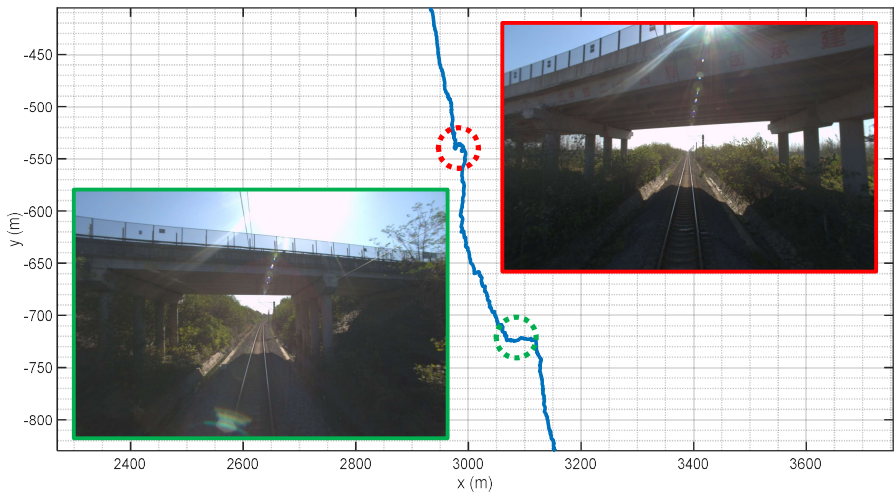

Figure 26. Examples of visual pose estimation failures caused by sudden light variations. The blue trajectory is the trajectory of VINS-Fusion, where the two images present the detailed scenes.

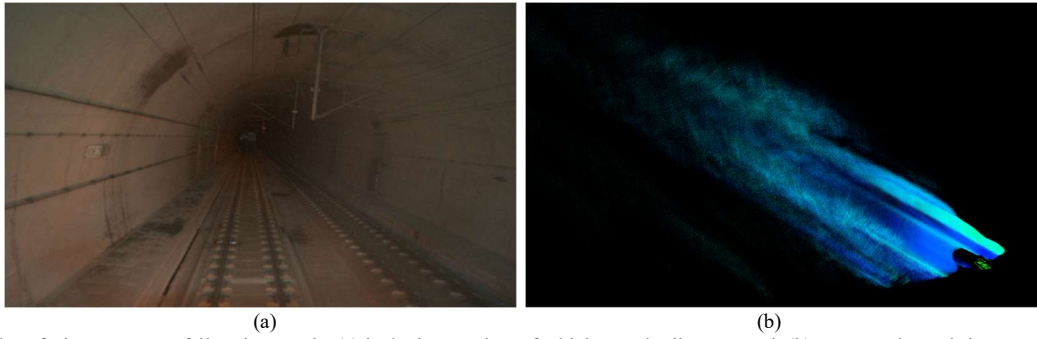

(a)                      (b)

Figure 27. Examples of LiDAR SLAM failure in tunnels. (a) is the image view of a high-speed railway tunnel. (b) presents the real-time mapping and odometry output of LOAM. It is seen the mapping is blurred with the odometry output 'stopped' due to degeneration.

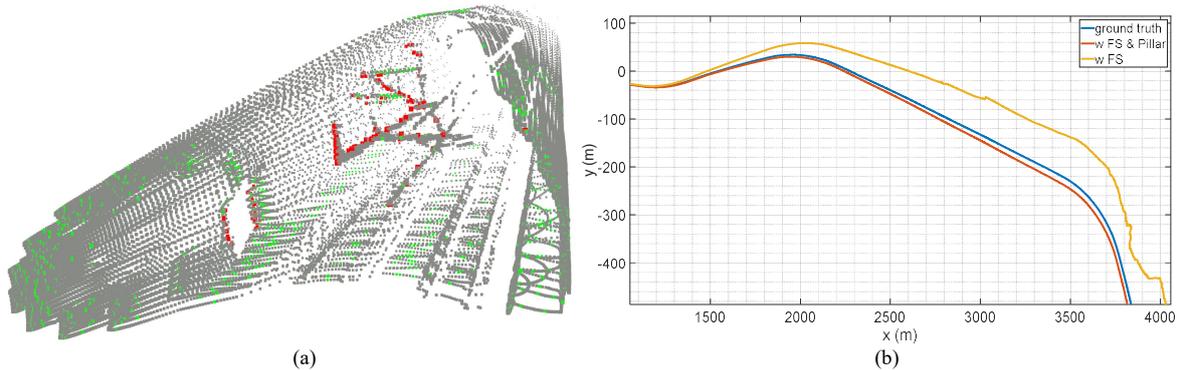

(a)                      (b)

Figure 28. (a) Visualization of the extracted feature points in high-speed railroad tunnels. The red points are the edge points, which mainly lie on overhead centenary systems and periodic cross passage tunnels. The green points indicate planar points, which are dominant in the tunnel environments. (b) Trajectory visualization in a tunnel of sequence *202109230014*. We denote w FS & Pillar as our method with feature selection and power pillar constraint, w FS is our method with merely feature selection method.

### 10.3.3 Robustness tests

In this section, we seek to evaluate our system robustness at typical scenes. Many visual-centric SLAM algorithms are sensitive to sudden illumination variations as pictured in Figure 26. As an inertial-centric SLAM, our algorithm is immune to this effect. For the LiDAR-inertial SLAM algorithms, we observe that the two rail tracks are not of the same height at turnings, and the LiDAR odometry will maintain this roll displacements even at the following straight railways. This happens frequently for LiDARs of limited FOV, and this error is not correctable even with GNSS measurements. The virtual track plane constraints now act as a strong correction for roll error compensation.

Tunnels is one of the most challenging scenarios for SLAM algorithms. With flat walls, no moving objects, and repetitive features such as rail tracks or power pillars, the railway tunnel is more difficult than many widely reported tunnel scenarios. While most of the state-of-the-art algorithms fails to provide meaningful results as visualized in Figure 27, our algorithm can still provide meter-level positioning accuracy in tunnels less than 4km. We believe this is mainly due to two reasons.

The first is removal of less informative features. We find that both the ICP-based or feature points based registration methods fail to track the frames constantly due to repetitive scenes. The planar point is the dominant features in the tunnel as visualized in Figure 28(a). Since most of them are similar and hard to track within frames, we adopt the feature selection methods to employ the most informative planar features for pose estimation. The second is finding changeable features in the repetitive scenes and make full use of such features. Although the rail tracks, walls, and power pillars are repetitive, the relative position of power pillars to the rail vehicles is changing continuously. We provide a comprehensive study using a 2.6 km tunnel of sequence *202109230014* in Figure 28(b). When neither feature selection method nor power pillar constraint is added for optimization, our method dies hundreds of meters inside the tunnel. With the assistance of feature selection, our method can still hold for several kilometers, but the heading angle cannot be estimated correctly.

### 10.3.4 Lessons learned

Lesson 1: A single estimation engine may not be robust for multi-modal system. We find that many unsynchronized messages or temporal loss of data exist throughout our test, and some LiDAR or visual centric SLAM may generate large pose estimation errors or die completely against these failures. This is mainly because of the single estimation design. Currently, many researches are focusing on developing estimation engines that incorporate all measurement jointly. However, the single estimation engine design may suffer from blocked buffer or information loss, and dies completely. Therefore, multiple estimation engines should be considered for highly reliable industrial applications.

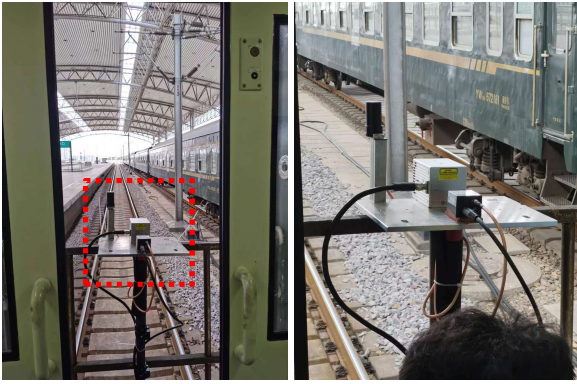

Figure 29. The sensor setup on the passenger train. The platform is placed at the caboose of each train. The system is composed of one Livox Avia LiDAR and one MTi-680G integrated navigation unit.

Lesson 2: Pay attention to variations in repetitive scenes to solve degeneration problem. Robust state estimation in featureless or repeated scenes is challenging. Instead of employing all features for calculation, we find that the changing objects contributes most to pose estimation. These variations including intensity and position differences. The intensity variations are the sudden change of point cloud reflectivity, such as the metal signs on the wall or the annunciators along the rail tracks. The objects that have relative position variations to us also provide strong constraint, including lamps, power pillars or road signs along the trajectory.

### 10.4 Tests with Passenger Train

Then we validate the multi-modal mapping and odometry on passenger train. Following the safety regulations on the railroad, the sensor platforms are sticked carefully at the caboose of the train. One Livox Avia and a MTi-680G is included in the system as shown in Figure 29. The data is captured and processed real-time by our customized onboard computer, with Intel i9 9980HK CPU, 64GB memory. The ground-truth is kept by the post-processing result of a MAPSTNAV M39 integrated navigation unit, with RTK corrections sent from Qianxun SI. Unlike the previous tests which only cover tens of kilometers, the passenger train journey is more than hundreds of kilometers. We conduct tests on train K8409 from Fuyang to Hefei, and the length is 221 km with the overall time consumption of 9317 s. The maximum speed limit for train K8409 is 120 km/h.

#### 10.4.1 Accuracy tests

We first evaluate the accuracy of our proposed multi-sensory system. For the first half journey, from Fuyang to Huainan (126 km, 4728 s), our algorithm runs real-time on the onboard computer while data gathering. Since no GPU is included in the computer, we disable the map visualization in ROS for fast response. The state estimation result is transformed into WGS-84 coordinates and projected onto satellite image as visualized in Figure 30. Besides, we also align two map segments with satellite image in Figure 30. The clear enough and well-aligned results demonstrate that our mapping is of high precision globally. In addition, we show some of the detailed mapping results both at low speeds and high speeds in Figure 31, which prove our mapping is clear even for large velocity.

Further, we conduct a more intensive test of the system accuracy. The 6 DoF errors along the path is pictured in Figure 32. We can see that our system maintains high positioning accuracy along the path, and the 3 DoF pose RMS is 1.38 m. Since the hardware is installed at the caboose of the train, the LiDAR facing is opposite to the moving direction. Fast-lio2 fails at the initialization stage as discussed in Section 10.2.2. Lili-om also fails at the high-speed stage while insufficient features are extracted.

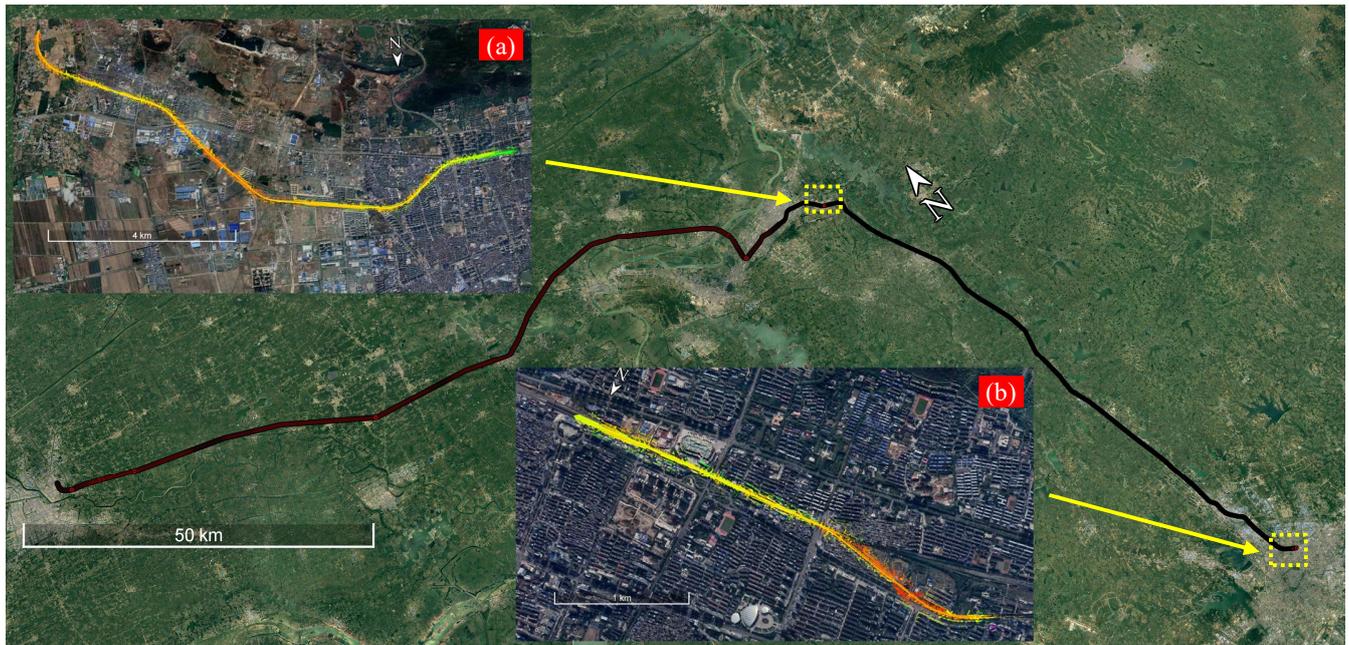

Figure 30. The multi-modal SLAM trajectory and detailed mapping plotted on the satellite image. The red line is the SLAM trajectory, (a) presents the map near Huainan station, (b) is the mapping near Hefei station.

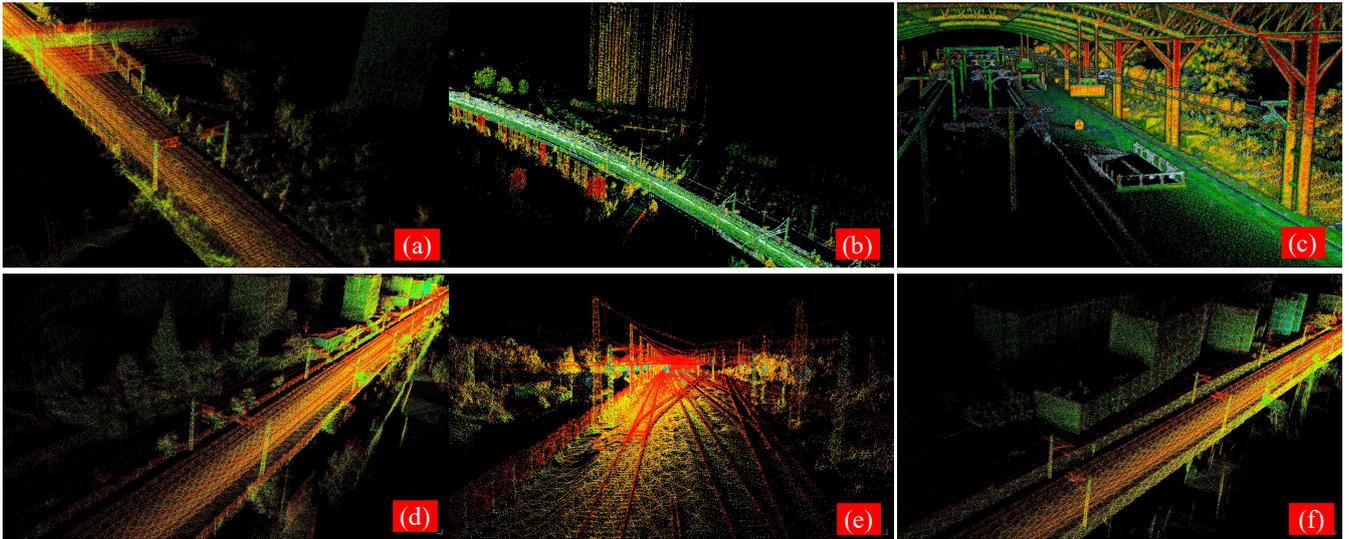

Figure 31. Some mapping details of passenger railway test. (a), (b), and (c) are the mapping at low speeds (0 to 60 km/h). (d), (e), and (f) are the results at high speeds (100 to 120 km/h).

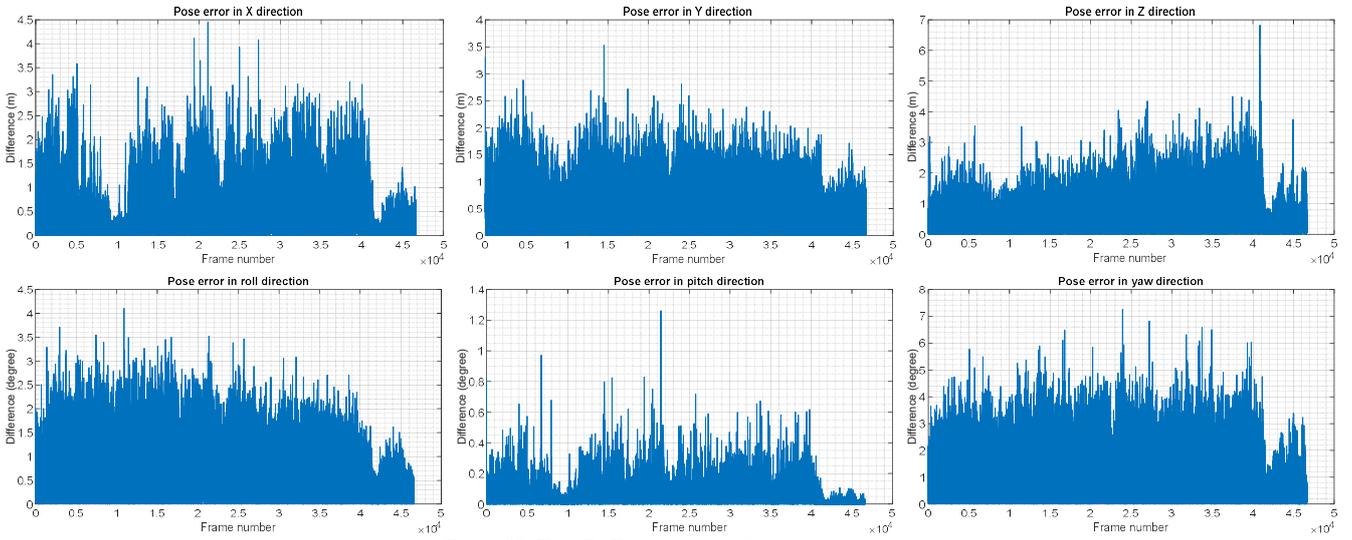

Figure 32. The 6DoF error curves of passenger train test.

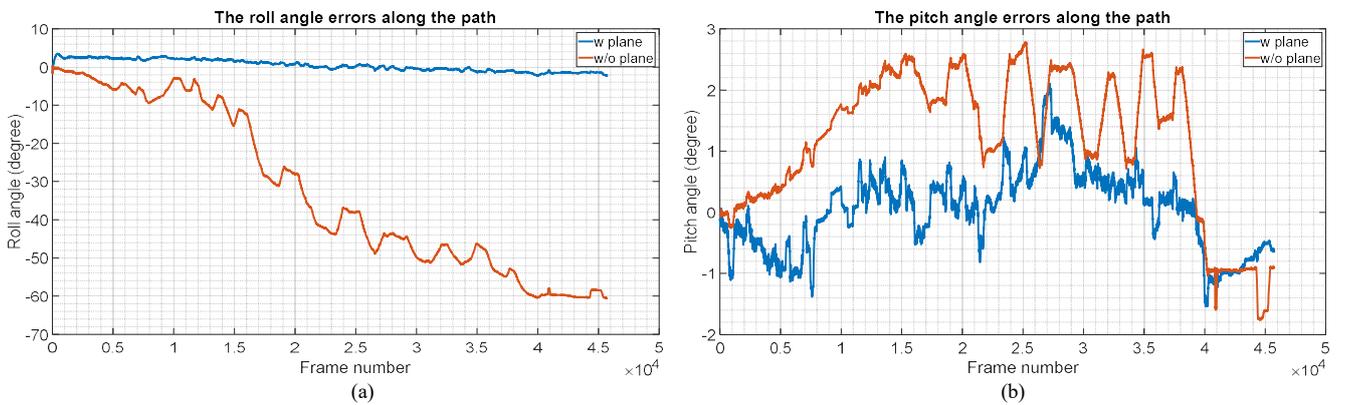

Figure 33. The roll and pitch angle difference of passenger train test.

### 10.4.2 Ablation study

We would like to further study the contribution of track plane for constraining horizontal roll and pitch errors. As discussed in Section 6, the LiDAR odometry of limited FoV LiDARs is prone to suffer from additional attitude errors. We denote w and w/o plane as our algorithm with or without track planes, and plot the error curves in Figure 33. Note that the GNSS optimization is enabled for both solutions. The results indicate that the roll directional error is not correctable and keeps increasing even with global optimization. This further prove that the extracted track plane can efficiently compensate for the roll directional errors. This is especially evident when the motion is aggressive. Although the global height measurements can compensate for the pitch errors, the local error is still much larger when plane constraint is not added.

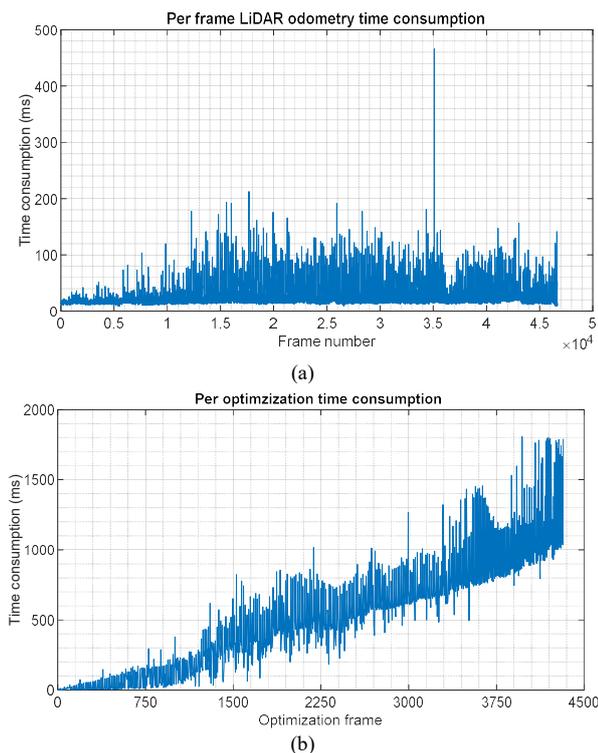

Figure 34. The time consumption statistics of LiDAR odometry and back-end optimization. (a) shows the per-frame LiDAR odometry time consumption and (b) presents the per-frame back-end optimization time consumption.

### 10.4.3 Runtime efficiency and resource consumption

Next, we would like to report the runtime efficiency and computation resource consumption of our system. As mentioned in Section 10.4.1, our system is running onboard for the first half of the journey, and we log the computation time of each module along the journey. The front-end process such as feature extraction and association and inertial preintegration has an almost constant time usage along the path. We plot the time consumption of LiDAR odometry in Figure 34(a), which is almost flat for the whole time. On the other hand, although the pose optimization problem is solved in the local window, the marginalized factors and map registration time is increased, and the global optimization time is in a rise. The back-end time usage is plotted in Figure 34(b), which is an incremental curve as discussed.

We also record the CPU and memory usage of our method along the journey. As visualized in Figure 35(a), the LiDAR odometry takes about 122.85% of a thread executing at 10 Hz. Considering the limited space, the cooling fan of the onboard computer is not well-designed, which lead to performance degradation for long-during operation. The processing time of LiDAR odometry thereby slightly increases along the journey. Similarly, this thread takes 0.60% to 1% memory. On the other hand, the map and pose optimization thread keeps occupying system resources. Although the CPU usage is half of that of LiDAR odometry at the beginning, it finally doubles the LiDAR odometry consumption. There are many sudden jumps for the thread memory usage curve, where maps are read into the memory for optimization.

### 10.4.4 Lessons learned

Lesson 1: The LiDAR odometry and mapping test on high-speed trains is currently not applicable. We have placed our LiDAR-inertial platform inside the driving cab of high-speed trains, but the result is not ideal. Currently, most of the cab windows are made of thick steel glass, where laser beams suffer from severe attenuation penetrating this glass. We observe that only 20% laser scans are received, non-sufficient for perception and localization. On the other hand, the platform cannot be placed anywhere outside the high-speed train, which will break up its aerodynamic configuration.

Lesson 2: For the LiDARs of restricted FoV, the attitude errors are still not avoidable even with GNSS optimizations. This is especially significant for roll directional divergence. We observe that although the x-y-z pose error is at a low level, the horizontal roll errors exceed 60 degrees for long-during tasks. Additional point cloud information such as plane constraints should be applied for compensation.

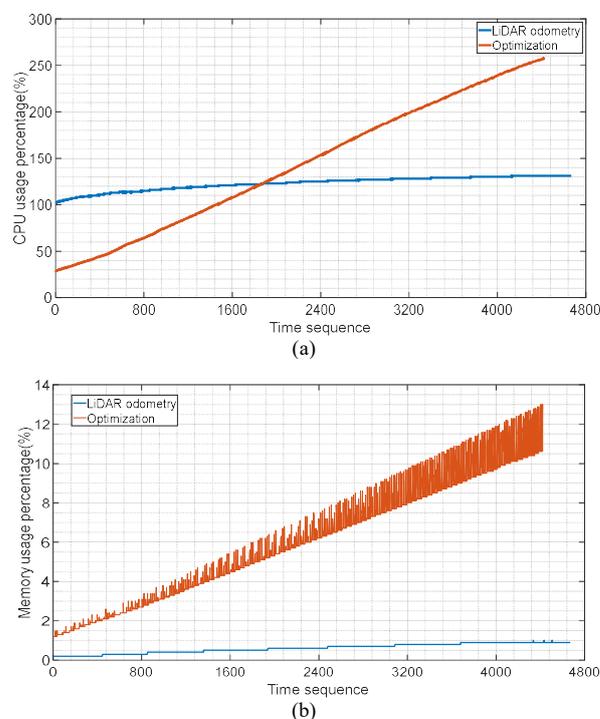

Figure 35. The per-second CPU and memory usage (in percentage) statistics. (a) is the CPU usage whereas (b) presents the memory usage.

## 10.5 Tests with autonomous driving rail vehicles

Our group have also conducted several automated maintenance rail vehicle experiments with Hefei University of Technology. They have developed an autonomous driving maintenance rail vehicle as shown in Figure 36, and our system can provide the train state information for localization and dispatch signal based navigation information. This test vehicle has a maximum cruise speed of 70 km/h to ensure the railway operation safety.

This experiment seeks to present the map-based localization performance without GNSS assistance. The automated vehicle is equipped with a Livox Horizon LiDAR and a camera. We store the base map on the server, and the onboard computer will automatically download the map sequence needed for matching. The ground truth trajectory is set by our previously integrated safety monitoring system (Lou et al., 2018), which is a combination of GNSS, IMU, odometer, and track map.

We employ two short sequences for illustration. The first is a slower one, with almost straight track lines, and the second has large velocity, with two curved track lines. The trajectory comparison as well as the absolute trajectory error (ATE) is plotted in Figure 37. It is seen that the map-based method has a vibrating curve at the initialization stage due to incorrect initial values. Considering the repetitive features in the direction of motion, the errors accumulate fast in this direction, and both the two end points are not coincided with the ground truth. Large velocity will lead to more point cloud distortion and we can see that the matching accuracy is worse in fast speed cases. Besides, multiple rail tracks can lead to horizontal matching errors.

We can infer that the map-based localization alone can achieve satisfactory performance on the railroads, and the error does not grow much with increased journey. The reason is twofold: First, the railroad environment is almost static, with seldom moving objects (trains) in the view. Second, the path of the maintenance rail vehicle is mostly straightforward. These factors determine the map matching is less likely to sudden lose tracking.

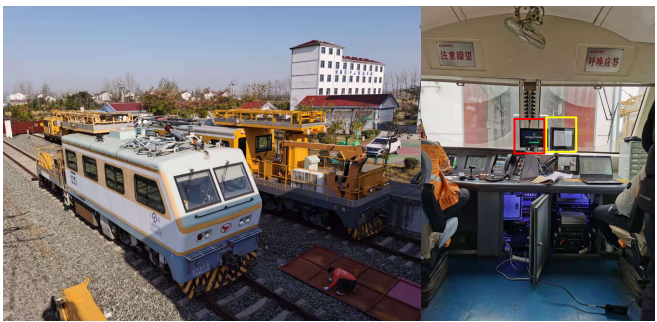

Figure 36. The exterior and interior of the autonomous driving maintenance rail vehicle. The screen in the red rectangle shows the real-time positioning result as well as the ATCS based warning information. The screen in the yellow rectangle shows the train automatic control information, such as pressure, brake, and acceleration.

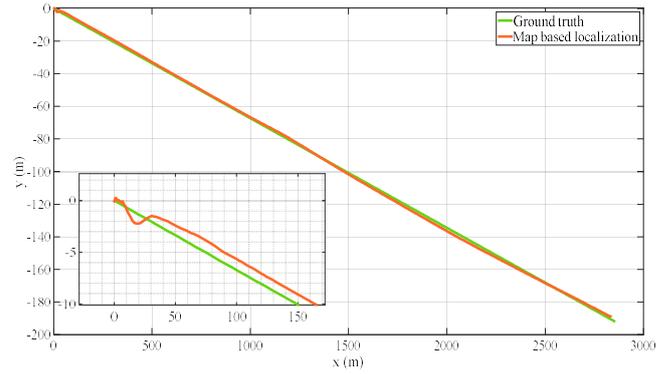

(a)

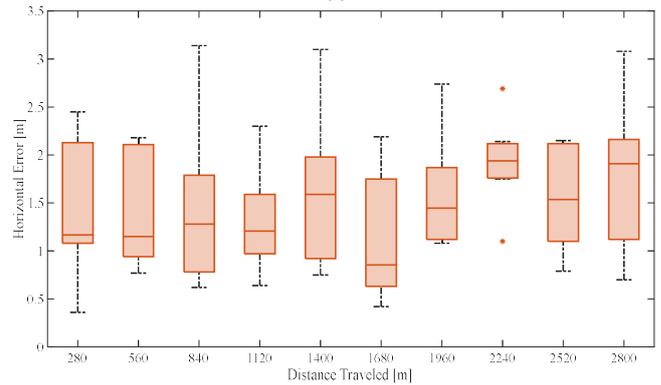

(b)

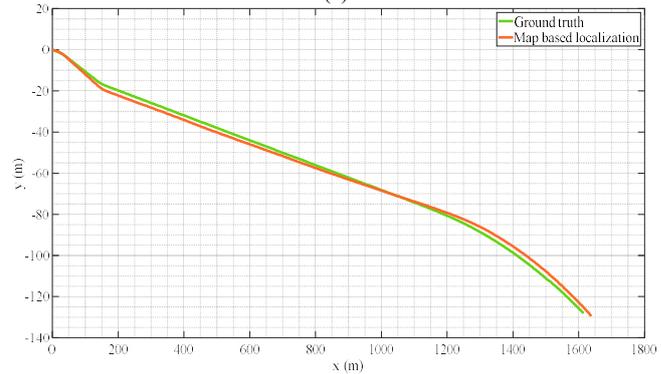

(c)

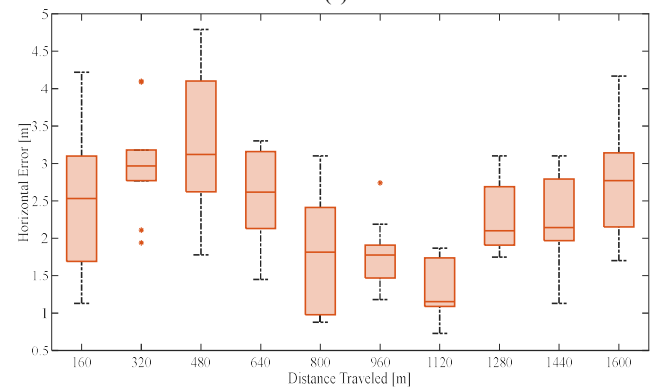

(d)

Figure 37. The accuracy evaluation of map-based localization method. (b) and (d) are the ATE of (a) and (c), respectively.

## 11 CONCLUSION

In this paper, our four-year work of multi-modal odometry and mapping for rail vehicles was presented. This environment contains a long list of distinctiveness that often make state-of-the-art methods fail. Our paper first offers an overview of the challenges and failure cases that many algorithms will probably encounter. Then we openly share our experiences in designing an accurate and robust framework for various rail vehicles. Our system is built atop an inertial-centric SLAM that is adaptive to a large variety of LiDAR or visual sensors. For the LiDAR-inertial odometry subsystem, we leverage constraints with geometric information to cope with the repeated environments. Similarly, we extract line segments in visual sequence to enhance the system accuracy and robustness. The proposed method has been extensively validated in large-scale railway, with a meter-level accuracy in most scenarios. The promising results for all platforms in different environments demonstrate that our system is of high robustness to individual sensor or estimation engine failures. In addition, our system has been successfully deployed for maintenance vehicle and railway environment monitoring.

There are two directions for future research. The first is to further improve the system robustness since the railroad requires super high reliability in all conditions. Integrating train control information into state estimation is desirable. Both the fusion of train state and the ATCS control data needs to be investigated.

Another direction regards engineering application, we seek to develop a full suite of localization and environment monitoring software. The current railroad environment monitoring and apparatus maintenance is still a human-intensive work. From our experience, several professional technicist need to go along with the maintenance vehicle every time to manually identify the fault spot, and this happens at least once a week. Although both human force and time consuming, this work still lacks effectiveness. If every manned and freight train is equipped with localization and perception sensors, the environmental changes can then be detected instantly. As described in Section 10.4, we have already conducted experiments on passenger train. We aim to develop a network that connects all the rail vehicles, where each train is equipped with localization and monitoring solution. While running, each train can report any abnormal cases to the governor, such as a tree that may fall on railroad. Then the governor will check this circumstance manually and decide whether to send a maintenance rail vehicle.

We hope that our experimental work and extensive evaluation could inspire follow-up works to explore multi-modal integration potential for rail applications. Besides, we wish the community pay more attention to railroad engineering applications, especially for facility and environment monitoring. A small advance towards autonomous system will save tremendous amount of manpower for railroad construction and maintenance.

ACKNOWLEDGMENT

This work was supported by the Joint Foundation for Ministry of Education of China under Grant 6141A0211907. We would like to thanks colleagues from Hefei power supply section, China Railway, for their kind support in data collection.

Video available at:
https://youtu.be/0VF2TrbE3go

Dataset available at:
https://github.com/YushengWHU/Railroad-dataset